%% file: main.tex
\documentclass[letterpaper]{article} 
\usepackage{aaai2027}  
\usepackage[hyphens]{url}  
\usepackage{graphicx} 
\graphicspath{{figures/}}
\usepackage{natbib}  
\usepackage{caption} 
\input{config}

\title{Probing Visual Idiosyncrasies in Text-to-Image Models}

\author {
    Yongqian Peng\textsuperscript{\rm 1,2}\equalcontrib,
    Hongjie Li\textsuperscript{\rm 3}\equalcontrib,
    Yichao Zhou\textsuperscript{\rm 2}\equalcontrib,
    Chi Zhang\textsuperscript{\rm 1,4}\corresponding,
    Yixin Zhu\textsuperscript{\rm 2,3,4,5,6,7}\corresponding
}
\affiliations {
    \textsuperscript{\rm 1}School of Intelligence Science and Technology, Peking University\\
    \textsuperscript{\rm 2}School of Psychological and Cognitive Sciences, Peking University\quad
    \textsuperscript{\rm 3}Institute for Artificial Intelligence, Peking University\\
    \textsuperscript{\rm 4}State Key Laboratory of General Artificial Intelligence, Peking University\\
    \textsuperscript{\rm 5} Beijing Key Laboratory of Brain-Computer Interface and Mental Health Modulation\\
    \textsuperscript{\rm 6} Embodied Intelligence Lab, PKU-Wuhan Institute for AI\quad
    \textsuperscript{\rm 7} XS Vision\\
    chizhang.cz@pku.edu.cn,
    yixin.zhu@pku.edu.cn
}

\begin{document}

\maketitle

\begin{abstract}
\ac{t2i} models turn short text descriptions into images, and the leading models now produce images of comparable visual quality.
Whether their outputs still carry model-specific characteristics matters for attribution, for model comparison beyond aggregate quality, and for evaluations that assume anonymity.
We ask whether images from frontier \ac{t2i} models can still be classified by source, and which visual properties make this possible.
Here we show that a standard ResNet-50 identifies the source of an image among \acl{hidream}, \acl{flux}, \acl{hunyuan}, and \acl{qwenimage}, the four highest-ranked open-source models on the Text-to-Image Arena, with 100.00\% accuracy on held-out prompts.
Human participants reach only 62.61\% on the pairwise version of this task, so the distinguishing cues are not visually obvious; they are also not tied to a particular classifier and survive severe image corruptions and out-of-distribution prompts.
We trace them to \textit{visual idiosyncrasies} at three levels: low-level texture statistics alone let a lightweight multi-layer perceptron match ResNet-50; object layout, depth, and albedo estimated from the images each carry a source signal; and image captions reveal model-specific preferences in style and background that the prompts never specify. A one-sentence hint about these cues raises human accuracy by 9.03 percentage points.
These idiosyncrasies make source classification a tool for measuring similarity between models and between real and generated images, and they caution against assuming anonymity in human preference evaluation of \ac{t2i} models.
Project page: \textcolor{magenta}{\url{https://ppyyqq.github.io/whichmodel/}}.
\end{abstract}

\acresetall
\section{Introduction}

\begin{figure}[t]
    \centering
    \includegraphics[width=\columnwidth]{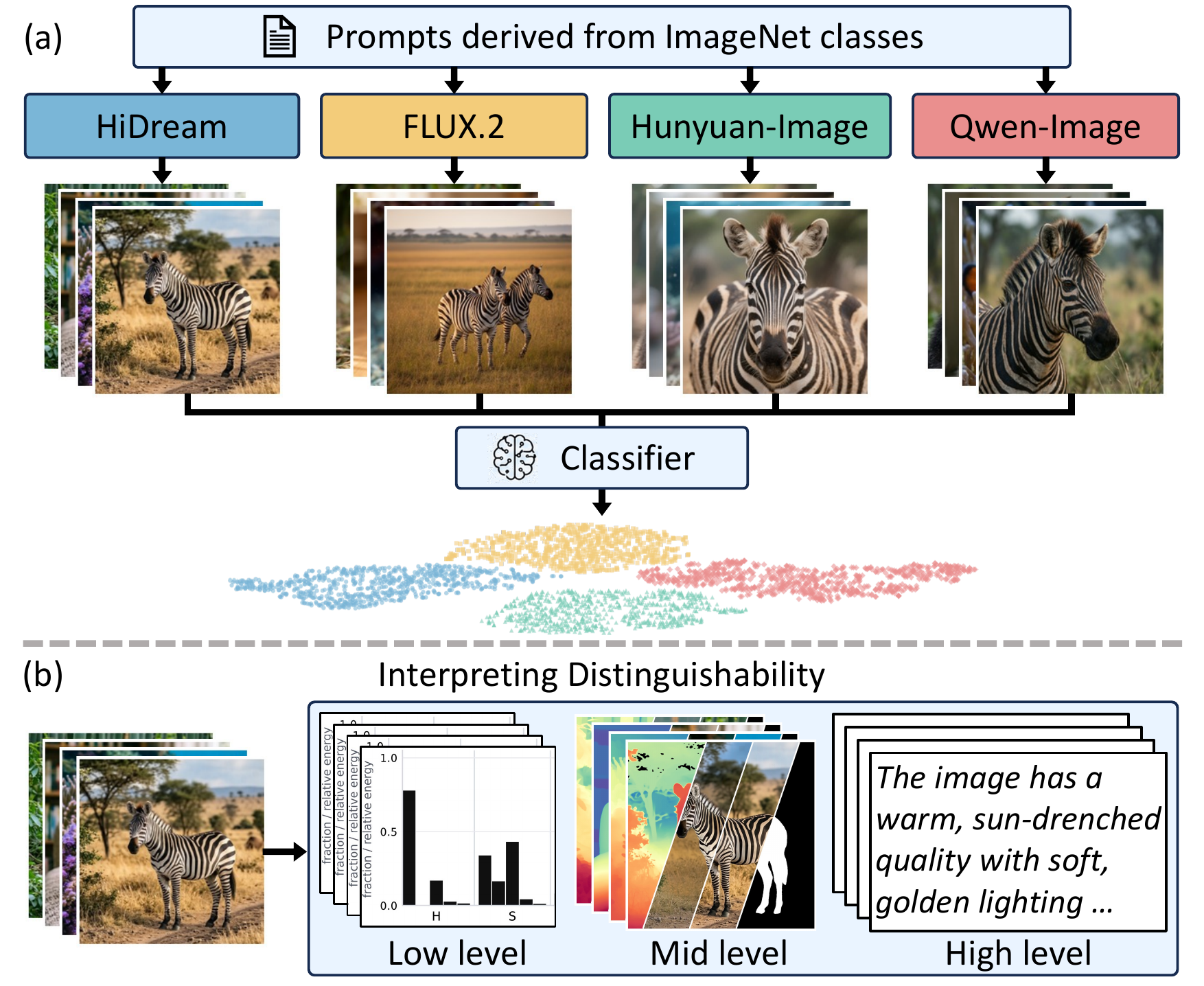}
    \caption{\textbf{Source classification and multi-level cue analysis.} (a) Images generated by four \acs{t2i} models from shared ImageNet-derived prompts are highly distinguishable by source under a trained classifier. (b) We further investigate this distinguishability by analyzing potential model-specific cues at low, mid, and high visual levels.}
    \label{fig:teaser}
\end{figure}

As \acf{t2i} models rapidly advance and produce images of increasingly comparable visual quality~\citep{arena2026texttoimage}, whether their outputs still carry model-specific characteristics becomes a pressing question. Prior work has shown that generated images can be attributed to their source models~\citep{girish2021towards,bui2022repmix,corvi2023detection,guarnera2024level}, and most recently that frontier \ac{t2i} sources remain distinguishable across models and generation settings~\citep{xu2025detecting}. These studies, however, treat attribution as the end goal and leave open what makes the sources distinguishable: the visual properties that systematically characterize each model remain largely unexplored. Characterizing these properties reveals how models differ beyond aggregate quality, supports model comparison and real-synthetic similarity analysis, and bears on evaluation protocols that assume model identities are hidden.

In this work, we ask whether outputs from frontier \ac{t2i} models can still be reliably classified by source, and which recurring visual properties enable this distinction; we call such properties \textit{visual idiosyncrasies}. We adopt a controlled source-classification task in the spirit of the ``Name That Dataset'' analyses of dataset bias~\citep{torralba2011unbiased,liu2025decade} and of recent work on idiosyncrasies of \acp{llm}~\citep{sun2025idiosyncrasies}: every model receives the same prompts, and a classifier must predict the source model from the image alone. We instantiate the task with four frontier open-source models (\ac{hidream}, \ac{flux}, \ac{hunyuan}, and \ac{qwenimage}) and dissect the cues that make their outputs distinguishable at the low, mid, and high levels of the visual hierarchy (\cref{fig:teaser}).

A standard ResNet-50~\citep{he2016deepresidual} identifies the source with 100.00\% accuracy on prompts held out from training, far above the 25\% chance level. This distinguishability is not a matter of visually obvious differences: on the pairwise version of the task, human participants average 62.61\%, whereas ResNet-50 averages 99.99\%. It is also not tied to a particular classifier, persists under severe image corruptions, and generalizes to \ac{ood} prompts.

We then characterize the visual idiosyncrasies behind this result at three levels. At the \textit{low level}, a bank of 225 handcrafted features suffices to distinguish the models, and a category-wise analysis singles out its 38 texture features: a lightweight \ac{mlp} trained on texture alone reaches 99.76\% four-way accuracy and generalizes to new prompts at least as well as ResNet-50. At the \textit{mid level}, object layout, depth, and albedo estimated from the images each carry a source signal, from 59.50\% accuracy for layout to 99.75\% for albedo, and the models differ systematically in average object shape, depth range, focal length, and the spatial scale of albedo variation. At the \textit{high level}, image captions expose preferences that the prompts never specify: using only \ac{tfidf} features of the captions, logistic regression classifiers reach up to 68.78\% and 71.78\% four-way accuracy from style and background descriptions, respectively.

These idiosyncrasies are useful beyond explaining the classifier. A one-sentence hint describing them raises human accuracy from 62.61\% to 71.64\%. Source classification also doubles as a similarity measure: leave-one-model-out prediction reveals which models resemble one another, and passing real ImageNet images through the classifiers at each level shows which model real photographs most resemble. Finally, persistent idiosyncrasies can betray a model's identity, which cautions against assuming anonymity in human preference evaluation such as the Text-to-Image Arena. Together, these results establish source classification as an interpretable probe of how frontier \ac{t2i} models differ, and visual idiosyncrasies as the properties it detects.

\section{Related Work}

\paragraph{Classification-based analysis of visual datasets.}
\citet{torralba2011unbiased} introduce the ``Name That Dataset'' experiment, in which a classifier that predicts which dataset an image comes from exposes dataset-specific bias. Such bias persists in modern large-scale datasets~\citep{liu2025decade} and arises from both structural and semantic factors~\citep{zeng2024understanding}. Our study is motivated by \citet{liu2025decade} and adopts the classification-based analyses of \citet{liu2025decade} and \citet{zeng2024understanding}, but targets model-specific rather than dataset-specific patterns: the classes are the source models of generated images from the same prompts, and the object of study is the visual properties that make them distinguishable.

\paragraph{Detection and attribution of AI-generated images.}
Synthetic images contain model-specific artifacts that enable detection~\citep{marra2019gans,chai2020what,wang2020cnn}; these artifacts take the form of spatial patterns, frequency signatures, textures, and style cues~\citep{liu2020global,frank2020leveraging,corvi2023intriguing,amoroso2024parents}. Recent detectors extend to diffusion models and generalize better with pretrained features~\citep{wang2023dire,ojha2023towards,cozzolino2024raising,liu2024forgery}, and zero-shot, training-free detectors exploit compression likelihood, autoencoder reconstruction, perturbation sensitivity, and diffusion-model geometry or dynamics~\citep{cozzolino2024zeroshot,ricker2024aeroblade,brokman2025manifold,choi2025trainingfree,liang2025denoising,choi2026debiased}. Attribution asks not whether an image is synthetic but which model produced it. \citet{yu2019attributing} show that GANs carry distinctive fingerprints, motivating attribution methods for GANs~\citep{girish2021towards,bui2022repmix} and for diffusion and other generative models~\citep{wang2023did,corvi2023detection,sha2023fake,wang2024latenttracer,laszkiewicz2024singlemodel,liu2024model,nguyen2025forensic,wang2026aedr}. Most closely related, \citet{xu2025detecting} study source distinguishability of \ac{t2i} models under varying models and settings, and \citet{naseh2026anonymity} deanonymize arena leaderboards through prompt-conditioned image-embedding clusters. Both establish that the source is recoverable; we ask what makes it recoverable and characterize the underlying visual idiosyncrasies at three levels.

\paragraph{Characterizing generative model idiosyncrasies.}
Beyond artifacts for detection and attribution, recent work characterizes properties of the generative models themselves. \citet{sarkar2024shadows} reveal systematic geometric inconsistencies in generated images, and \citet{lee2026csf} show that underspecified prompts expose model-specific semantic preferences. In the language domain, \citet{sun2025idiosyncrasies} use controlled classification to identify idiosyncrasies across frontier \acp{llm}. We bring this classification-based lens to images and characterize the idiosyncrasies of frontier \ac{t2i} models across low-, mid-, and high-level representations.

\section{Probing Visual Distinguishability of\\\texorpdfstring{\ac{t2i}}{T2I} Models}

\paragraph{Formulation.}
We formulate source classification as a controlled task. Let $\mathcal{M}=\{m_1,\ldots,m_N\}$ denote a set of \ac{t2i} models and $\mathcal{P}$ a shared prompt set. For each prompt $p\in\mathcal{P}$ and model $m_i\in\mathcal{M}$, we sample an image $x\sim m_i(\cdot\mid p)$ and assign it the source label $y=i$. A classifier observes only $x$ and predicts $y$. Because every model receives the same prompts, prompt content cannot explain above-chance accuracy; accuracy above the balanced chance level of $1/N$ therefore indicates visual properties that consistently characterize a model across prompts, which we term \emph{visual idiosyncrasies}.

\paragraph{Experimental setup.}
We instantiate the task with four recent \ac{t2i} models: \ac{hidream}~\citep{cai2026hidream}, \ac{flux}~\citep{blackforestlabs2025flux2}, \ac{hunyuan}~\citep{cao2025hunyuanimage}, and \ac{qwenimage}~\citep{qwen2025qwenimage2512}. As of June 2026, they are the four highest-ranked open-source models on the Text-to-Image Arena~\citep{arena2026texttoimage}, each in the overall top 40. For each model, we generate images from prompts derived from the 1,000 ImageNet classes~\citep{russakovsky2015imagenet}, with four random seeds per prompt, yielding 4,000 images per model and 16,000 in total. Images are generated at $1024\times1024$ with each model's default sampling configuration (examples in \cref{fig:appx_image_examples}). We hold out 200 classes as the \ac{id} test set and the remaining classes as training classes. Our default classifier is a ResNet-50~\citep{he2016deepresidual} trained from random initialization on images resized to $224\times224$. Prompt construction and training details are in \cref{sec:appx_prompts,sec:appx_classifiers}.

\begin{table}[t!]
    \centering
    \small
    \setlength{\tabcolsep}{3.5pt}
    \begin{tabular}{ccccc}
        \toprule
        \acs{hidream} & \acs{flux} & \acs{hunyuan} & \acs{qwenimage} & Acc. \\
        \midrule
        \cmark & \cmark & & & 100.00 \\
        \cmark & & \cmark & & 100.00 \\
        \cmark & & & \cmark & 99.94 \\
        & \cmark & \cmark & & 100.00 \\
        & \cmark & & \cmark & 100.00 \\
        & & \cmark & \cmark & 100.00 \\
        \midrule
        \cmark & \cmark & \cmark & \cmark & 100.00 \\
        \bottomrule
    \end{tabular}
     \caption{\textbf{Source classification accuracy across model combinations.} ResNet-50 accuracy (\%) on the held-out ImageNet-derived test set for all six pairwise tasks and the four-way task. Every combination is nearly perfectly distinguishable.}
    \label{tbl:source_combinations}
\end{table}

\begin{table}[t]
    \centering
    \small
    \setlength{\tabcolsep}{11pt}
    \begin{tabular}{lcc}
        \toprule
        Classifier & Acc. & Acc. (Random labels) \\
        \midrule
        ResNet-50 & 100.00 & 26.41 \\
        Two-layer \acs{cnn} & 98.81 & 26.63 \\
        DINOv2 ViT-B/14 & 88.54 & 25.94 \\
        CLIP ViT-B/32 & 90.77 & 24.97 \\
        \bottomrule
    \end{tabular}
    \caption{\textbf{Source classification across classifier architectures.} Four-way accuracy (\%) on the held-out ImageNet-derived test set, and accuracy after training on randomly permuted labels. All classifiers are far above the 25\% chance level, and random-label performance remains near chance.}
    \label{tbl:classifier_architectures}
\end{table}

\subsection{High Distinguishability Across \texorpdfstring{\ac{t2i}}{T2I} Models}

\paragraph{Consistency across source combinations and classifiers.}
\Cref{tbl:source_combinations} reports ResNet-50 accuracy for all pairwise tasks and the four-way task. Every pair is perfectly distinguishable; the hardest, \ac{hidream} versus \ac{qwenimage}, still reaches 99.94\%. Expanding the label space to all models barely matters: the four-way classifier reaches 100.00\%. This distinguishability is not specific to one architecture. A two-layer \ac{cnn} trained from scratch and linear classifiers on frozen DINOv2 ViT-B/14~\citep{oquab2023dinov2} and CLIP ViT-B/32~\citep{radford2021clip} features differ widely in capacity, inductive bias, and pretraining objective, yet all remain far above chance (\cref{tbl:classifier_architectures}; confusion matrices in \cref{sec:appx_confusion_main}). Nor does it stem from arbitrary correlations in the training labels: with randomly permuted labels, every classifier drops to the 25\% chance level (\cref{tbl:classifier_architectures}, right column).

\begin{figure}[t]
    \centering
    \includegraphics[width=\columnwidth]{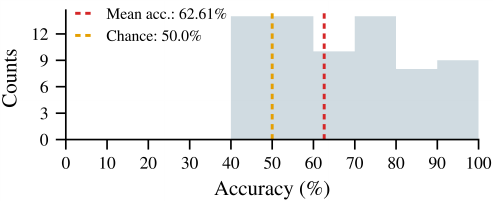}
    \caption{\textbf{Human accuracy on pairwise source classification.} Participants ($n=69$) average 62.61\% with wide individual variation, far below the 99.99\% that ResNet-50 averages on the same model pairs.}
    \label{fig:human_accuracy_histogram}
\end{figure}

\begin{figure}[t]
    \centering
    \includegraphics[width=\columnwidth]{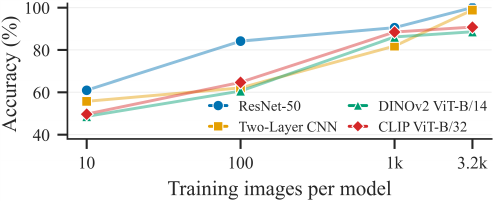}
    \caption{\textbf{Accuracy as a function of training set size.} Accuracy on the held-out test set rises steadily with the number of training images per model, and 10 images per model already give non-trivial accuracy.}
    \label{fig:training_size_scaling}
\end{figure}

\begin{table}[t]
    \centering
    \small
    \setlength{\tabcolsep}{14.5pt}
    \begin{tabular}{lcc}
        \toprule
        Corruption & Setting & Acc. \\
        \midrule
        None & - & 100.00 \\
        \midrule
        Color jittering & Strength: 1.0 & 100.00 \\
        Color jittering & Strength: 2.0 & 99.94 \\
        Gaussian blur & Radius: 4.0 & 100.00 \\
        Gaussian blur & Radius: 8.0 & 100.00 \\
        Gaussian noise & Blending weight: 0.4 & 99.91 \\
        Gaussian noise & Blending weight: 0.8 & 97.99 \\
        Low resolution & Resolution: 32$\times$32 & 100.00 \\
        Low resolution & Resolution: 8$\times$8 & 99.21 \\
        Low resolution & Resolution: 2$\times$2 & 84.74 \\
        \bottomrule
    \end{tabular}
    \caption{\textbf{Robustness to image corruptions.} ResNet-50 accuracy (\%) when training and test images are subjected to the same corruption. Accuracy stays nearly perfect under moderate settings and well above chance under the most severe ones; the full sweeps are in \cref{tbl:appx_corruption_full}.}
    \label{tbl:corruption_robustness}
\end{table}

\begin{figure}[t]
    \centering
    \includegraphics[width=\columnwidth]{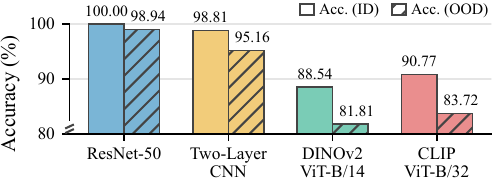}
    \caption{\textbf{Cross-prompt generalization.} Classifiers are trained on images generated from ImageNet-derived prompts and evaluated on images generated from Winoground-derived prompts (\acs{ood}). All four remain far above the 25\% chance level.}
    \label{fig:ood_prompt_generalization}
\end{figure}

\paragraph{Model differences are not fully perceptible to humans.}
To test whether this distinguishability rests on visually obvious differences, we conduct a two-alternative forced-choice study on Prolific. Each of 69 participants is assigned to one of the six model pairs, studies a reference board of nine images per model labeled only as Set~A and Set~B, and then classifies ten new images by source (\cref{sec:appx_human_study}). Participants average 62.61\% accuracy, far below the 99.99\% that ResNet-50 averages over the same pairwise tasks, and vary widely (\cref{fig:human_accuracy_histogram}). Visually obvious differences alone therefore cannot account for the distinguishability: the classifiers capture model-specific patterns more consistently than humans do.

\subsection{Distinguishability Persists Across Conditions}

\paragraph{Accuracy improves with more training data.}
We vary the number of training images and retrain all classifiers (\cref{fig:training_size_scaling}). Accuracy on the held-out test set rises steadily with training size, and even 10 images per model give non-trivial accuracy (\eg, 60.96\% for ResNet-50), well above chance.

\paragraph{Robustness to image corruptions.}
To test robustness to input degradation, we next train and evaluate ResNet-50 on images subjected to one of four corruptions: color jitter~\citep{krizhevsky2012imagenet}, Gaussian blur, blending with Gaussian noise, and resolution reduction. Classification stays nearly perfect under moderate settings and degrades only under severe ones, yet remains well above chance even at a noise blending weight of 0.8 or a resolution of $2\times2$ (\cref{tbl:corruption_robustness}; examples in \cref{fig:appx_corruption_examples}). Two further transformations show that the signal is neither purely global nor a fragile decoder artifact: shuffling $32\times32$ patches leaves accuracy at 99.43\%, and re-encoding every image through an off-the-shelf \ac{vae} leaves it above 99\% (\cref{sec:appx_corruptions}).

\paragraph{Generalization to out-of-distribution prompts.}
Finally, we train on ImageNet-derived prompts and evaluate on 1,600 images per model generated from the 400 Winoground prompts~\citep{thrush2022winoground}, which describe compositional relations among multiple entities rather than single object categories. All classifiers remain far above chance, and ResNet-50 retains 98.94\% (\cref{fig:ood_prompt_generalization}). The distinguishability thus extends beyond the training prompt distribution to a different kind of visual content.

\section{Characterizing Visual Idiosyncrasies of\\\texorpdfstring{\ac{t2i}}{T2I} Models}

Having established that classifiers identify the source model reliably, we ask which model-specific properties make this possible. We analyze three complementary levels: low-level visual cues, mid-level visual representations, and high-level semantic realization. Qualitative examples for all three analyses are collected in \cref{sec:appx_examples}.

\subsection{Low-Level Visual Cues}

\paragraph{Low-level visual cues distinguish \texorpdfstring{\ac{t2i}}{T2I} models.}
We construct a bank of 225 handcrafted features in seven categories, computed from each native $1024\times1024$ image: (i) \textbf{tone} (luminance statistics, dynamic range, entropy, and local contrast); (ii) \textbf{color} (RGB and perceptual chromaticity statistics, color histograms, inter-channel correlations, and colorfulness); (iii) \textbf{texture} (co-occurrence statistics, local binary patterns, and local entropy); (iv) \textbf{edge and sharpness} (gradient-orientation distributions, edge density, and high-frequency sharpness); (v) \textbf{frequency} (spectral energy distributions and oriented frequency responses); (vi) \textbf{spatial organization} (regional luminance and edge distributions, symmetry, and luminance centroids); and (vii) \textbf{residual and noise} (high-pass residual statistics, spatial autocorrelation, and residual spectral energy). The composition and definitions are given in \cref{sec:appx_feature_bank,sec:appx_feature_definitions}. A lightweight \ac{mlp} trained on the full bank (\cref{sec:appx_low_level_classifier}) reaches 99.84\% accuracy on the \ac{id} test set and 98.55\% on the \ac{ood} set (\cref{tbl:feature_mlp}). Low-level statistics alone thus support reliable source attribution across prompt distributions, with OOD accuracy close to ResNet-50’s 98.94\% (\cref{fig:ood_prompt_generalization})

\begin{table}[t]
    \centering
    \small
    \setlength{\tabcolsep}{9pt}
    \resizebox{\columnwidth}{!}{%
    \begin{tabular}{lcccc}
        \toprule
        \multirow{2}{*}{Feature category}
        & \multicolumn{2}{c}{Acc. (Single)}
        & \multicolumn{2}{c}{Acc. (Perm.)} \\
        \cmidrule(lr){2-3} \cmidrule(lr){4-5}
        & \acs{id} & \acs{ood} & \acs{id} & \acs{ood} \\
        \midrule
        Full
        & 99.84 & 98.55
        & - & - \\
        \midrule
        Tone
        & 91.27 & 85.18
        & 98.85 & 94.88 \\
        Color
        & 93.28 & 91.10
        & 96.36 & 92.93 \\
        Texture
        & \textbf{99.76} & \textbf{99.07}
        & \textit{89.51} & \textit{88.46} \\
        Edge and sharpness
        & 94.41 & 89.96
        & 99.68 & 98.09 \\
        Frequency
        & 99.67 & 98.36
        & 99.31 & 97.42 \\
        Spatial organization
        & 95.20 & 90.91
        & 99.20 & 97.14 \\
        Residual and noise
        & 99.33 & 97.60
        & 96.60 & 93.46 \\
        \bottomrule
    \end{tabular}%
    }   
      \caption{\textbf{Source classification with handcrafted features.}
      Accuracy (\%) of an \acs{mlp} trained on the full feature bank
      (top row) and on each category alone (Single), and accuracy of
      the full-bank \acs{mlp} after permuting one category across test images (Perm.), on the held-out ImageNet-derived (\acs{id}) and Winoground-derived (\acs{ood}) sets. Among individual feature categories, texture achieves the highest single-category accuracy (bold) and the largest accuracy decrease under grouped permutation (italic).}
      \label{tbl:feature_mlp}
  \end{table}

\begin{figure}[t]
    \centering
    \includegraphics[width=\columnwidth]{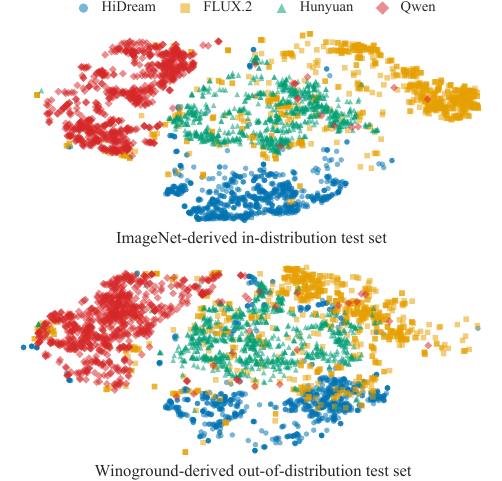}
    \caption{\textbf{t-SNE embedding of the 38 texture features.} A joint embedding of 3,200 images from each test set, shown separately for the ImageNet-derived and Winoground-derived sets. The model-dependent structure persists across the two prompt distributions.}
    \label{fig:low_level_texture_tsne}
\end{figure}

\paragraph{Texture features capture persistent model differences.}
We assess each category's standalone predictive power and grouped
permutation importance with two experiments
(\cref{sec:appx_category_analysis}): training a separate \ac{mlp}
on one category alone, and permuting one category across test images
while keeping the trained full-bank \ac{mlp} and the other features
fixed. Texture stands out on both measures
(\cref{tbl:feature_mlp}). Its 38 features alone reach 99.76\% and
99.07\% on the \ac{id} and \ac{ood} sets, respectively. Permuting
texture reduces full-bank accuracy to 89.51\% and 88.46\%, producing
the largest decrease among all feature categories, about ten
percentage points on both sets. Frequency and residual features
are also individually informative (99.67\% and 99.33\% on the
\ac{id} set), but their permutation causes smaller accuracy
decreases, at most 5.1 percentage points. The texture category comprises eight \ac{glcm} statistics~\citep{haralick1973textural}, two uniform \ac{lbp} histograms~\citep{ojala2002multiresolution} with $(P,R)=(8,1)$ and $(16,2)$, and local-entropy statistics. A joint t-SNE embedding~\citep{van2008visualizing} of these features for 3,200 images from each test set shows clear model-dependent structure that persists across the two prompt distributions (\cref{fig:low_level_texture_tsne}). Texture is thus a prominent source of visual idiosyncrasies; visualizations of its four components and of the other six categories are in \cref{sec:appx_low_level_viz,sec:appx_texture_stats}.

\subsection{Mid-Level Visual Representations}

\paragraph{Mid-level representations show model differences.}
We test whether the differences survive when an image is reduced to a mid-level representation. We estimate three: (i) \textbf{object layout} as segmentation masks from SAM~3.1~\citep{carion2026sam}; (ii) \textbf{geometry} as depth maps from MoGe-2~\citep{wang2025moge}; and (iii) \textbf{appearance} as albedo maps from DiffusionRenderer~\citep{liang2025diffusion} (\cref{sec:appx_mid_level_extraction}; examples in \cref{fig:appx_mid_level_examples}). We train and evaluate a separate ResNet-50 on each representation. All three remain above chance on the \ac{id} set and lose only moderately on the \ac{ood} set (\cref{tbl:mid_level_representation}): even binary object masks reach 59.50\%, depth reaches 87.72\%, and albedo 99.75\% (confusion matrices in \cref{sec:appx_mid_level_confusion}). The same ordering holds for the other classifier architectures (\cref{sec:appx_mid_level_classifiers}).

\begin{table}[t]
    \centering
    \small
    \setlength{\tabcolsep}{14pt}
    \begin{tabular}{lcc}
        \toprule
        Representation & Acc. (\acs{id}) & Acc. (\acs{ood}) \\
        \midrule
        Segmentation mask & 59.50 & 46.73 \\
        Depth map & 87.72 & 70.50 \\
        Albedo map & 99.75 & 97.73 \\
        \bottomrule
    \end{tabular}
    \caption{\textbf{Source classification from mid-level representations.} Accuracy (\%) of a separate ResNet-50 trained on each estimated representation. All three remain above the 25\% chance level on both test sets, and albedo is nearly as informative as the original image.}
    \label{tbl:mid_level_representation}
\end{table}

\begin{figure}[t]
    \centering
    \includegraphics[width=\columnwidth]{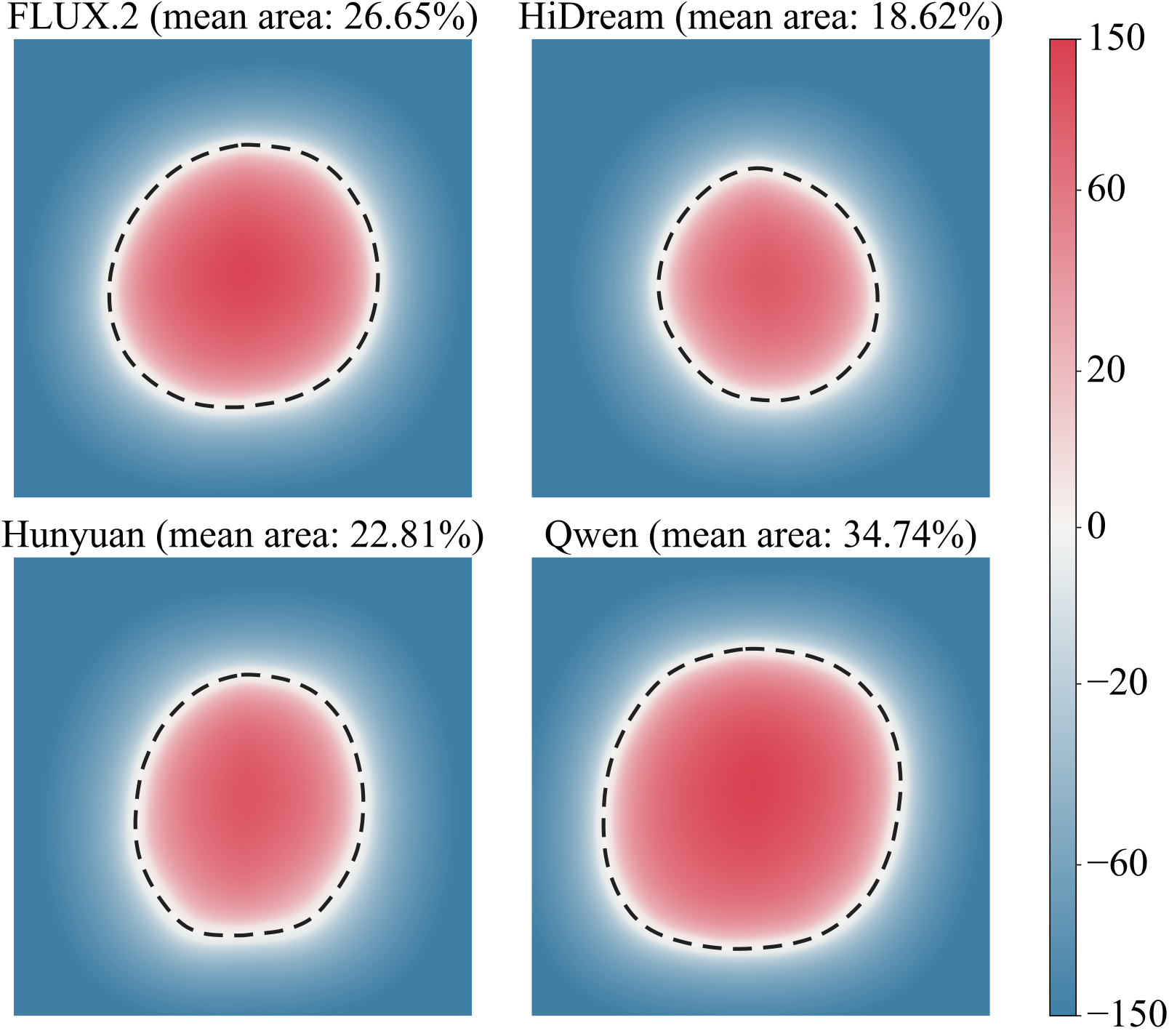}
    \caption{\textbf{Average object layouts.} Signed distance maps (in pixels) derived from segmentation masks on the ImageNet-derived set, averaged per model; dashed lines mark the zero-level contours. \acs{qwenimage} has the largest average object.}
    \label{fig:average_layout}
\end{figure}

\begin{figure}[t]
    \centering
    \includegraphics[width=\columnwidth]{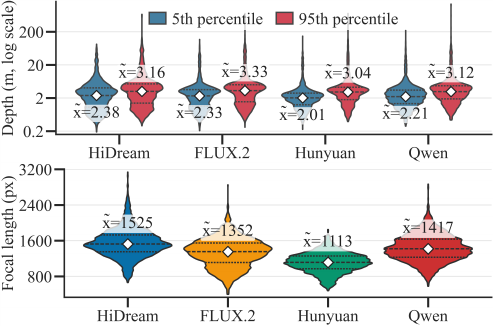}
    \caption{\textbf{Geometry differences.} Per-image 5th and 95th percentiles of object depth and the focal lengths estimated by MoGe-2. \acs{hunyuan} combines the nearest objects with the shortest focal lengths.}
    \label{fig:geometry_violin_plot}
\end{figure}

\begin{figure}[t]
    \centering
    \includegraphics[width=\columnwidth]{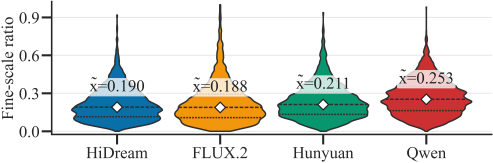}
    \caption{\textbf{Appearance differences.} Fine-scale variation ratio of the albedo map, the share of albedo variation that occurs at fine spatial scales; higher values indicate more fine-grained albedo changes.}
    \label{fig:appearance_violin_plot}
\end{figure}

\paragraph{Decoding model-specific mid-level visual structure.}
The representations also reveal what differs. For layout, we convert each mask into a signed distance map and average the maps per model over matched images (\cref{sec:appx_mid_level_statistics}); the averaged layouts differ visibly, and \ac{qwenimage} in particular renders a substantially larger average object than the other models (\cref{fig:average_layout}). For geometry, the models occupy distinct ranges of object depth, summarized by the 5th and 95th depth percentiles of each image (\cref{fig:geometry_violin_plot}). Depth interacts with the camera, so we also compare the focal lengths estimated by MoGe-2: \ac{hunyuan} has markedly shorter focal lengths, \ie, a wider field of view, which explains how its objects can be both nearer in depth and smaller in the image. For appearance, we measure the albedo map's fine-scale variation ratio, the share of albedo variation at fine rather than coarse spatial scales (\cref{sec:appx_mid_level_statistics}). \ac{qwenimage} has the highest median fine-scale variation ratio (0.253), while \ac{hidream} and \ac{flux} have similarly lower medians (0.190 and 0.188), indicating more spatially coherent albedo regions. \ac{hunyuan} is intermediate, with a median of 0.211 (\cref{fig:appearance_violin_plot}). Mid-level representations thus offer a second, interpretable window on the visual idiosyncrasies of \ac{t2i} models.

\subsection{High-Level Semantic Realization}

\begin{table}[t]
    \centering
    \small
    \setlength{\tabcolsep}{9pt}
    \begin{tabular}{llcc}
        \toprule
        Captioner & Prompt & Acc. (\acs{id}) & Acc. (\acs{ood}) \\
        \midrule
        Qwen3-VL & \multirow{3}{*}{Style} & 65.62 & 60.16 \\
        InternVL3.5 & & 66.06 & 54.78 \\
        GLM-4.1V & & 68.78 & 58.19 \\
        \midrule
        Qwen3-VL & \multirow{3}{*}{Background} & 71.12 & 65.11 \\
        InternVL3.5 & & 70.94 & 60.53 \\
        GLM-4.1V & & 71.78 & 62.47 \\
        \bottomrule
    \end{tabular}
    \caption{\textbf{Source classification from \acs{vlm}-generated captions.} Accuracy (\%) of logistic-regression classifiers on \acs{tfidf} features of style and background captions from three captioners; chance is 25\%. Classification stays above chance for every captioner and both prompt distributions.}
    \label{tbl:high_level_semantics}
\end{table}

\begin{table}[t]
    \centering
    \small
    \setlength{\tabcolsep}{2pt}
    \resizebox{\columnwidth}{!}{%
        \begin{tabular}{llll}
            \toprule
            \acs{hidream} & \acs{flux} & \hunyuanstack & \qwenimagestack \\
            \midrule
            \textbf{window} & \textbf{warm} & hazy & \textbf{dark} \\
            natural light & showing \textbf{warm} & \textbf{sunlit} & green \\
            cloudy sky & golden & blurred with & is \textbf{dark} \\
            brick & beige & bokeh & muted \\
            partly cloudy & pale & weathered & green foliage \\
            \midrule
            \textbf{window} & \textbf{warm} & sunlight & green \\
            people & \textbf{warm} neutral & dusty & \textbf{dark} \\
            brick & paved area & \textbf{sunlit} & doorway \\
            walking & golden & branches suggesting & featuring green \\
            focus setting & features light & soft & is \textbf{dark} \\
            \midrule
            \textbf{window} & \textbf{warm} & \textbf{sunlit} & \textbf{dark} \\
            \textbf{window} with & with \textbf{warm} & atmosphere & green \\
            brick & smooth & hazy & is \textbf{dark} \\
            flowers & neutral & soft & \textbf{dark} and \\
            wooden & grassy area & muted & muted \\
            \bottomrule
        \end{tabular}
    }
    \caption{\textbf{Model-specific background preferences revealed by captions.} The five unigrams or bigrams with the largest positive coefficients of a word-level \acs{tfidf} logistic-regression classifier for background descriptions; the three row blocks correspond, from top to bottom, to Qwen3-VL, InternVL3.5, and GLM-4.1V. Bold words rank among the top terms for the same model under all three captioners.}
    \label{tbl:high_level_semantics_word}
\end{table}

\paragraph{High-level semantic realization reveals preferences of \texorpdfstring{\ac{t2i}}{T2I} models.}
Since all models receive identical prompts, any difference in how they realize the same concept reflects a preference of the model. We examine two aspects that the prompts leave unspecified, \textbf{visual style} and \textbf{background context}, by asking three \ac{vlm} captioners, Qwen3-VL~\citep{bai2025qwen3}, InternVL3.5~\citep{wang2025internvl3}, and GLM-4.1V~\citep{hong2025glm}, to describe each image's style and background (\cref{sec:appx_captioning}; example captions in \cref{tbl:appx_caption_examples}). We represent each caption by word- and character-level \ac{tfidf} features and train a logistic-regression classifier to predict the source model (\cref{sec:appx_caption_classifier}). Classification stays well above chance for both aspects and all three captioners, on the \ac{id} and \ac{ood} sets alike (\cref{tbl:high_level_semantics}; confusion matrices in \cref{sec:appx_high_level_confusion}), so the models differ systematically in how they realize style and background.

\paragraph{Semantic realization patterns are model-specific.}
To see what differs, we fit a word-only \ac{tfidf} classifier per captioner and list, for each model, the five unigrams or bigrams with the largest positive coefficients for background descriptions (\cref{tbl:high_level_semantics_word}; style in \cref{sec:appx_style_words}). Several associations recur across all three captioners. \ac{hidream} backgrounds are associated with window and brick, \ac{flux} with warm, \ac{hunyuan} with sunlit, and \ac{qwenimage} with dark and green. These recurring terms characterize source-associated background realization under shared prompts. Individual coefficients describe predictive associations within the four-source roster between terms and sources.

\section{Applications and Implications}

We illustrate three applications and one implication of the visual idiosyncrasies beyond explaining the classifier.

\begin{figure}[t]
    \centering
    \includegraphics[width=\columnwidth]{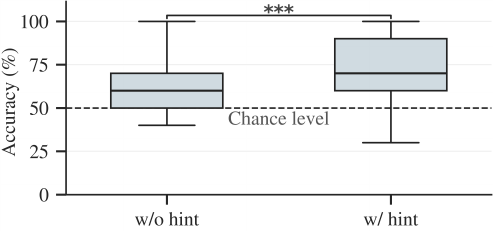}
    \caption{\textbf{Effect of a hint on human source classification.} Asking participants to compare style, texture, and background raises mean accuracy from 62.61\% ($n=69$) to 71.64\% ($n=61$). Statistical significance: $^{*}p<0.001$ (stratified permutation test).}
    \label{fig:human_study_hint}
\end{figure}

\paragraph{A probe for guiding human judgment.}
We test whether the identified idiosyncrasies help people tell models apart. We repeat the human study of \cref{fig:human_accuracy_histogram} with 61 new participants from the same pool and change only one thing: every question adds the hint ``Please focus on comparing the images in terms of style, texture, and background,'' which summarizes the cues found at the low and high levels. The hint raises mean accuracy from 62.61\% to 71.64\% (\cref{fig:human_study_hint}); the 9.03-point gain is significant under a stratified permutation test ($p<0.001$) and positive in five of the six model pairs (\cref{sec:appx_human_stats}). The idiosyncrasies thus serve as interpretable probes that make model differences visible to people.

\paragraph{Quantifying inter-model visual similarity.}
We assess visual similarity by comparing sources through leave-one-source-out prediction. We exclude one source, train ResNet-50 on the other three, and report its predictions for the excluded source (\cref{tbl:model_similarity}). Held-out \ac{hidream} is predominantly assigned to \ac{flux} (93.34\%). Held-out \ac{flux} is assigned to \ac{hidream} (58.29\%) and \ac{hunyuan} (31.28\%). \ac{hunyuan} divides almost equally between \ac{flux} (49.37\%) and \ac{hidream} (48.49\%), while \ac{qwenimage} is most often assigned to \ac{hidream} (76.88\%). These projections describe how each unseen source is partitioned by a classifier trained on the remaining sources.

\begin{table}[t]
    \centering
    \small
    \setlength{\tabcolsep}{4pt}
    \begin{tabular}{lcccc}
        \toprule
        Held-out model & \acs{hidream} & \acs{flux} & \hunyuanstack & \qwenimagestack \\
        \midrule
        \acs{hidream} & - & \textbf{93.34} & 3.02 & \underline{3.64} \\
        \acs{flux} & \textbf{58.29} & - & \underline{31.28} & 10.43 \\
        \acs{hunyuan} & \underline{48.49} & \textbf{49.37} & - & 2.14 \\
        \acs{qwenimage} & \textbf{76.88} & \underline{20.23} & 2.89 & - \\
        \bottomrule
    \end{tabular}
    \caption{\textbf{Inter-model similarity from leave-one-model-out classification.} Each row gives the distribution (\%) of sources predicted for the held-out model's images by a ResNet-50 trained on the other three models. Bold marks the most frequent prediction and underline the second.}
    \label{tbl:model_similarity}
\end{table}

\begin{table}[t]
    \centering
    \small
    \setlength{\tabcolsep}{3pt}
    \begin{tabular}{lcccc}
        \toprule
        Representation & \acs{hidream} & \acs{flux} & \hunyuanstack & \qwenimagestack \\
        \midrule
        Original & \underline{7.37} & \textbf{89.16} & 1.90 & 1.56 \\
        \midrule
        Handcrafted feature & \textbf{47.09} & \underline{42.12} & 1.41 & 9.38 \\
        \midrule
        Segmentation mask & 21.66 & 21.48 & \textbf{30.48} & \underline{26.38} \\
        Depth map & \underline{26.53} & \textbf{36.87} & 14.15 & 22.45 \\
        Albedo map & \underline{26.68} & \textbf{54.69} & 10.12 & 8.51 \\
        \midrule
        Caption (style) & 20.88 & \textbf{36.11} & 13.63 & \underline{29.37} \\
        Caption (background) & 13.22 & \textbf{39.55} & 16.19 & \underline{31.04} \\
        \bottomrule
    \end{tabular} 
    \caption{\textbf{Mean predicted source probabilities (\%) for real ImageNet images across representation levels.} Each row averages the probabilities produced by the source classifier trained on the corresponding representation. Bold marks the largest probability and underline the second largest.}
    \label{tbl:real_synthetic_similarity}
\end{table}

\paragraph{Assessing real-synthetic similarity.}
We evaluate the trained source classifiers on 10,000 real ImageNet images, using the corresponding input representation for each classifier (\cref{sec:appx_real_images}). \cref{tbl:real_synthetic_similarity} reports mean predicted probabilities. The RGB classifier predominantly assigns probability to \ac{flux} (89.16\%), while the handcrafted-feature classifier favors \ac{hidream} (47.09\%) followed by \ac{flux} (42.12\%). Mask predictions are more dispersed, with \ac{hunyuan} receiving 30.48\%. Depth, albedo, style captions, and background captions favor \ac{flux}. The responses differ across representation levels and characterize the behavior of these source-trained classifiers on real images.

\paragraph{Risks to anonymous evaluation.}
Persistent visual idiosyncrasies challenge the anonymity assumed by human preference evaluations such as the Text-to-Image Arena~\citep{arena2026texttoimage}. Even when model identities are hidden, the images themselves may reveal the source, and a participant can use a trained source classifier to infer a generator and vote for or against that identity. Our human study addresses a separate question: whether a hint about the characterized cues improves human source identification; it is not an empirical test of an arena attack. \citet{naseh2026anonymity} demonstrate leaderboard deanonymization through prompt-conditioned image-embedding clusters; our results complement this attack-oriented view by characterizing, for more recent models, the multi-level cues that make deanonymization possible.

\section{Conclusion}

Images from four frontier \ac{t2i} models remain almost perfectly distinguishable by source under identical prompts, across classifier architectures, under severe corruptions, and on out-of-distribution prompts, even though humans distinguish them only modestly. This robustness suggests that the distinguishing signal reflects stable properties of each model's learned output distribution rather than superficial artifacts of prompting, resolution, or post-processing. We trace this distinguishability to visual idiosyncrasies at three levels: texture statistics that alone suffice for classification, systematic differences in object layout, depth, focal length, and albedo, and model-specific style and background preferences that captions expose. Together, these findings indicate that frontier models with similar benchmark scores still occupy noticeably different regions of image space. These idiosyncrasies make source classification an interpretable probe of how generative models differ, guide human judgment, quantify inter-model and real-synthetic similarity, and warn against assuming anonymity in model evaluation. Limitations and future directions are discussed in \cref{sec:appx_limitations}.

\section*{Acknowledgments}
This work is supported in part by the: Brain Science and Brain-like Intelligence Technology-National Science and Technology Major Project (2026ZD0222200), the National Natural Science Foundation of China (62376009), the Beijing Nova program, the NVIDIA Academic Grant Program using Spark and Thor, the PKU-BingJi Joint Laboratory for Artificial Intelligence, the Wuhan Major Scientific and Technological Special Program (2025060902020304), the Hubei Embodied Intelligence Foundation Model Research and Development Program, and the National Comprehensive Experimental Base for Governance of Intelligent Society, Wuhan East Lake High-Tech Development Zone.

\bibliography{aaai2027}

\setcounter{secnumdepth}{2}
\clearpage
\acresetall
\appendix
\renewcommand\thefigure{A\arabic{figure}}
\setcounter{figure}{0}
\renewcommand\thetable{A\arabic{table}}
\setcounter{table}{0}
\renewcommand\theequation{A\arabic{equation}}
\setcounter{equation}{0}
\pagenumbering{arabic}
\renewcommand*{\thepage}{A\arabic{page}}
\setcounter{footnote}{0}

\section{Experimental Details}\label{sec:appx_details}

\subsection{Prompt Construction}\label{sec:appx_prompts}

All source models receive the same prompts, so that prompt content is controlled. We use two prompt families: ImageNet-derived prompts, which name a single object category, and Winoground-derived prompts, which describe compositional relations among multiple entities. Example images for both families are shown in \cref{fig:appx_image_examples}.

\paragraph{ImageNet-derived prompts.}
We construct one prompt for each of the 1,000 ImageNet-1K classes~\citep{russakovsky2015imagenet}. For a class label with a primary name followed by comma-separated aliases, the primary name follows ``a photo of'' and the aliases are appended after ``which is also known as''; the second clause is omitted when a class has no alias. Four seeds per prompt yield 4,000 images per source model.

\promptbox{ImageNet-derived prompt template and examples}{%
    \textit{Template:} \texttt{a photo of \{primary class name\}}\\
    \hspace*{3.7em}\texttt{[, which is also known as \{alias 1\}, \{alias 2\}, \ldots]}\\[3pt]
    \textit{Examples:}\\
    \texttt{a photo of stingray}\\
    \texttt{a photo of tabby, which is also known as tabby cat}\\
    \texttt{a photo of goldfish, which is also known as Carassius auratus}\\
    \texttt{a photo of toilet tissue, which is also known as toilet paper, bathroom tissue}%
}

\paragraph{Winoground-derived prompts.}
Winoground~\citep{thrush2022winoground} contains 400 examples, each a contrastive pair $(\texttt{caption\_0}, \texttt{caption\_1})$ that reverses or alters a compositional relation. We use \texttt{caption\_0} verbatim as the generation prompt, without any photographic or stylistic wrapper; \texttt{caption\_1} is not used. The same four seeds yield 1,600 images per source model, which form the \ac{ood} test set.

\promptbox{Winoground-derived prompt template and examples}{%
    \textit{Template:} \texttt{\{caption\_0\}}\\[3pt]
    \textit{Examples:}\\
    \texttt{an old person kisses a young person}\\
    \texttt{a bird eats a snake}\\
    \texttt{a bottle is in water}\\
    \texttt{light colored dog with dark colored cat}%
}

\begin{figure}[t]
    \centering
    \includegraphics[width=\columnwidth]{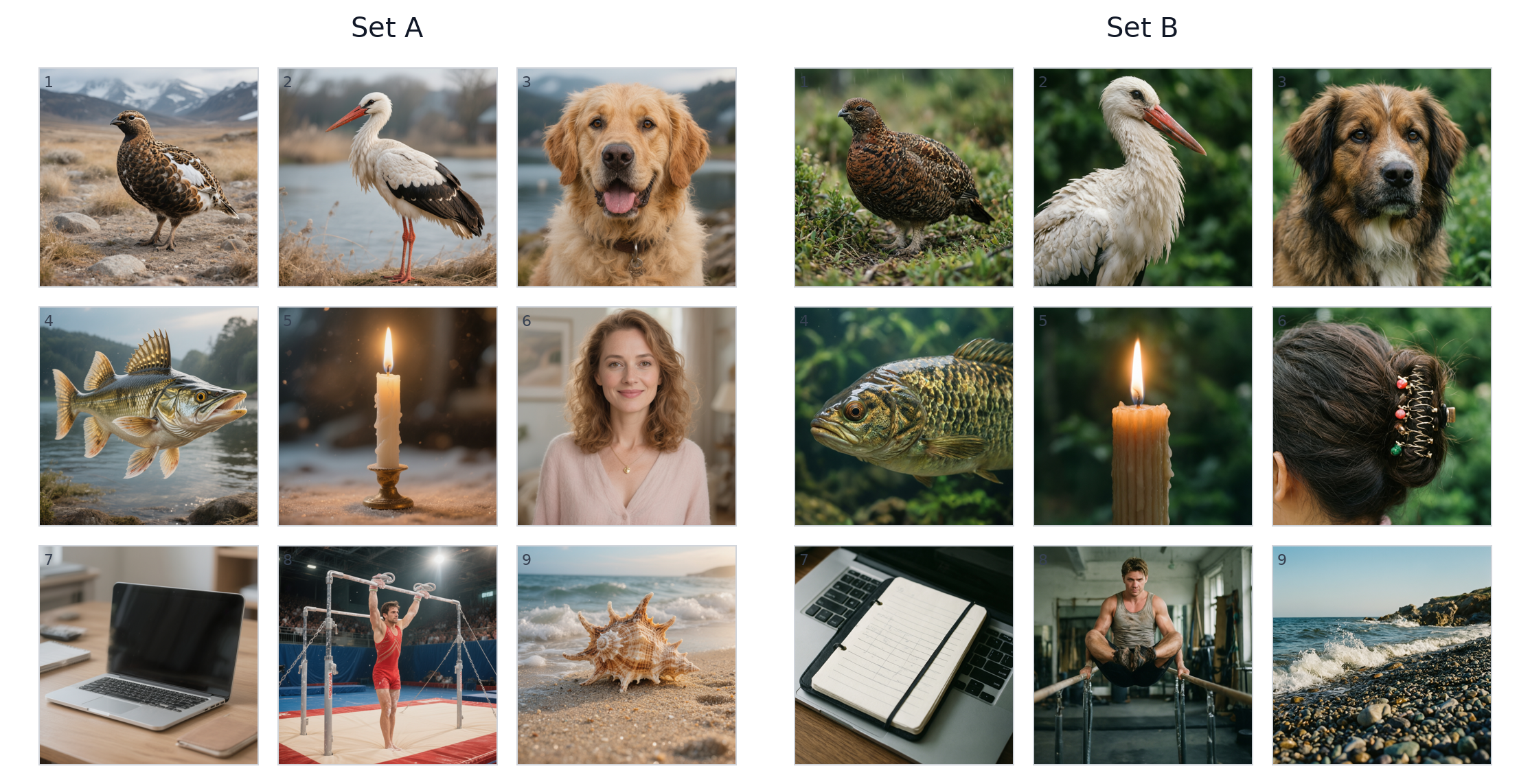}
    \caption{\textbf{Example reference board used in the human study.} Each anonymous set contains nine examples from one source model, with matched grid positions generated from the same prompt. The assignment of models to Set~A and Set~B is counterbalanced across study versions.}
    \label{fig:appx_reference_board}
\end{figure}

\begin{figure*}[t]
    \centering
    \begin{subfigure}[t]{0.4\textwidth}
        \centering
        \includegraphics[width=\linewidth]{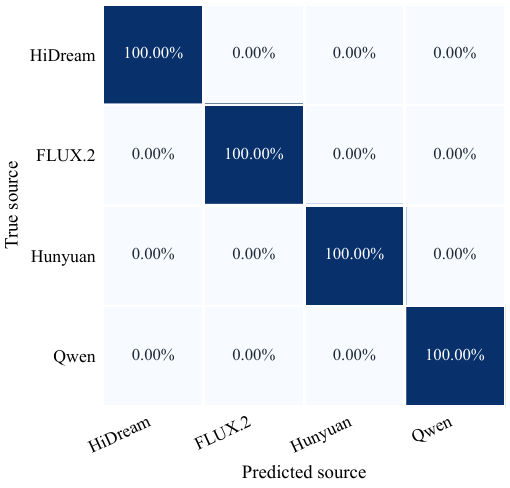}
        \caption{ResNet-50.}
    \end{subfigure}
    \qquad
    \begin{subfigure}[t]{0.4\textwidth}
        \centering
        \includegraphics[width=\linewidth]{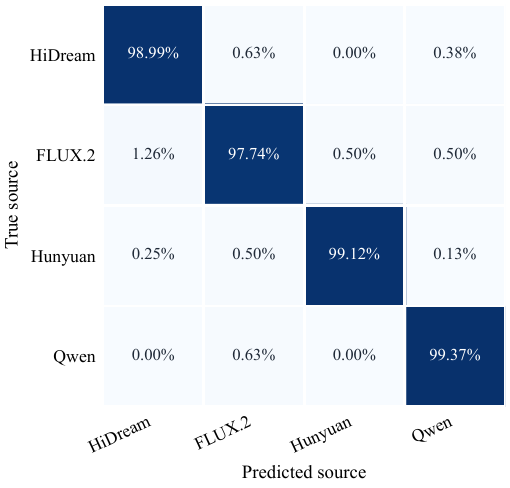}
        \caption{Two-layer \acs{cnn}.}
    \end{subfigure}

    \begin{subfigure}[t]{0.4\textwidth}
        \centering
        \includegraphics[width=\linewidth]{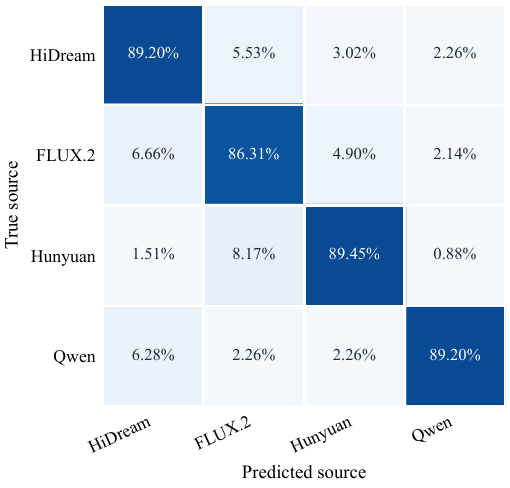}
        \caption{Linear classifier on DINOv2 features.}
    \end{subfigure}
    \qquad
    \begin{subfigure}[t]{0.4\textwidth}
        \centering
        \includegraphics[width=\linewidth]{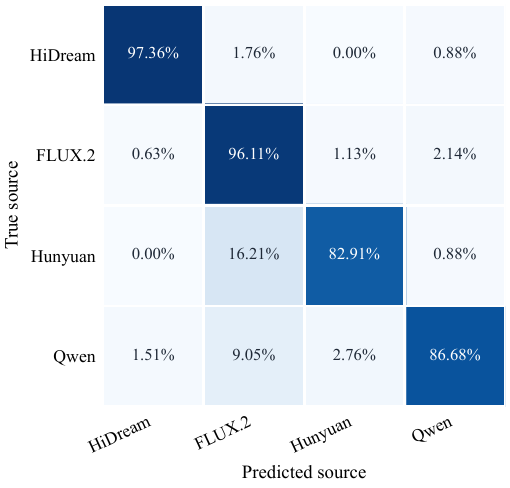}
        \caption{Linear classifier on CLIP features.}
    \end{subfigure}
    \caption{\textbf{Source-classification confusion matrices on the held-out ImageNet-derived test set.} Rows denote true source models and columns denote predicted sources. Each cell reports the row-normalized percentage.}
    \label{fig:appx_confusion_main}
\end{figure*}

\subsection{Source Classifiers and Training}\label{sec:appx_classifiers}

\paragraph{Architectures.}
ResNet-50~\citep{he2016deepresidual} is trained from random initialization with its final fully connected layer replaced by an $N$-way classification head. The two-layer \ac{cnn} uses two $3\times3$ convolutional layers with 32 and 64 output channels, each followed by ReLU and $2\times2$ max pooling; the resulting feature map is adaptively average-pooled to $8\times8$, flattened, and passed through a two-layer head with dimensions $4096\rightarrow128\rightarrow N$ and an intermediate ReLU. For DINOv2 ViT-B/14~\citep{oquab2023dinov2} and CLIP ViT-B/32~\citep{radford2021clip}, we freeze the visual encoders and train a single linear classifier on their image features.

\paragraph{Data splits.}
We partition the 1,000 ImageNet-derived classes into 800 training classes (200--999) and 200 \ac{id} evaluation classes (0--199). With four generation seeds and four source models, this yields 12,800 training and 3,200 \ac{id} evaluation images, balanced across sources. No separate validation split is used: checkpoints are selected by accuracy on the training set. Feature normalization is fitted only on the training set. All 6,400 Winoground-derived images form the \ac{ood} test set and are used only for final evaluation, without affecting normalization or checkpoint selection.

\paragraph{Training.}
Each image is decoded, converted to RGB, and resized to $224\times224$ with bicubic interpolation. ResNet-50 and two-layer \ac{cnn} inputs are normalized channel-wise with mean 0.5 and standard deviation 0.5; DINOv2 and CLIP inputs use their respective model-specific preprocessing. We apply no stochastic data augmentation. All classifiers are trained with cross-entropy loss and Adam~\citep{adam} at a fixed learning rate of $10^{-4}$ for 100 epochs with a batch size of 16. ResNet-50 and the two-layer \ac{cnn} are optimized end-to-end, whereas the DINOv2 and CLIP encoders remain frozen. We select the checkpoint with the highest training accuracy, and report its accuracy on the held-out test sets. All experiments run on a single NVIDIA H100 GPU.

\subsection{Human Study Protocol}\label{sec:appx_human_study}

\paragraph{Participants and design.}
We recruit adult participants through Prolific and administer the study in English using Google Forms. Participants provide informed consent, confirm that they are at least 18 years old and hold at least a bachelor's degree, and are instructed to use a desktop or laptop computer.

The four source models form six unordered model pairs. Each participant is assigned to one pair and learns to distinguish its two sources, anonymously labeled Set~A and Set~B. For each pair, we counterbalance which model is assigned to which set and cross this mapping with two instruction conditions, with and without a hint, yielding $6\times2\times2=24$ between-participant study versions. Repeat participation is prevented through Prolific. New participants are automatically assigned to one of the currently least-filled versions, and returning participants are directed back to their original version.

We receive 150 submissions and exclude participants who answer either of two attention checks incorrectly, leaving 130 participants: 69 in the no-hint condition, analyzed in the main text, and 61 in the hint condition. All retained participants complete the ten main trials and both attention checks.

\paragraph{Stimuli.}
All stimuli are sampled reproducibly from the ImageNet-derived set. The same nine reference prompts and ten main-trial prompts are used across all model pairs, with no overlap between reference and trial prompts. Within each pair, the two models are always compared on images generated from matched prompts. Each reference board shows nine examples from each source as two $3\times3$ grids, with matched grid positions generated from the same prompt (\cref{fig:appx_reference_board}). The two mapping versions use identical images and differ only in which source is labeled Set~A. The ten main trials follow a fixed prompt order and contain five query images from each source.

\paragraph{Procedure.}
Participants complete one trial per page. Each page displays the assigned reference board, a query image with its prompt, and the question \textit{Which reference set is this image more likely to come from?} In the hint condition, the sentence \textit{Hint: Please focus on comparing the images in terms of style, texture, and background.} is added to every question. Participants select Set~A or Set~B; model identities and correct answers are never shown. The sequence consists of three main trials, one attention check, three main trials, one attention check, and four main trials. Each attention check reuses an image from the reference board and is presented as an ordinary trial. The primary outcome is participant-level mean accuracy over the ten main trials, with chance at 50\%; attention checks and a final cue-ranking question are excluded from this computation.
\begin{table}[t]
    \centering
    \small
    \setlength{\tabcolsep}{7pt}
    \begin{tabular}{lcc}
        \toprule
        Setting \((s,g)\) & \ac{flux} recall & Overall acc. \\
        \midrule
        \((50,4)\) & 99.75 & 99.94 \\
        \((25,4)\) & 99.87 & 99.97 \\
        \((75,4)\) & 99.87 & 99.97 \\
        \((50,2)\) & 99.75 & 99.94 \\
        \((50,6)\) & 99.87 & 99.97 \\
        \bottomrule
    \end{tabular}
    \caption{\textbf{Source attribution across \ac{flux} generation settings.}
    Here \(s\) is the number of inference steps and \(g\) is distilled guidance.
    A single frozen four-way ResNet-50 evaluates 800 \ac{flux} images per setting.
    Overall accuracy also includes 800 matched images from each of \ac{hidream}, \ac{hunyuan}, and \ac{qwenimage}.}
    \label{tbl:appx_flux_settings}
\end{table}
\begin{table}[t]
    \centering
    \small
    \setlength{\tabcolsep}{19.5pt}
    \begin{tabular}{lcc}
        \toprule
        Comparison & Centroid & \(\mathrm{MMD}^{2}\) \\
        \midrule
        \ac{flux} \((25,4)\) & 0.0148 & 0.000167 \\
        \ac{flux} \((75,4)\) & 0.0076 & 0.000045 \\
        \ac{flux} \((50,2)\) & 0.0248 & 0.000469 \\
        \ac{flux} \((50,6)\) & 0.0136 & 0.000144 \\
        \midrule
        \ac{hidream} & 0.1688 & 0.021128 \\
        \ac{hunyuan} & 0.1587 & 0.018803 \\
        \ac{qwenimage} & 0.1722 & 0.022227 \\
        \bottomrule
    \end{tabular}
    \caption{\textbf{CLIP distances within \ac{flux} and across generators.}
    Every row compares the indicated group with the newly generated \ac{flux} \((50,4)\) reference.
    The four \ac{flux} rows vary generation settings, and the final three rows compare different generators.
    Each group contains 800 images matched by prompt and generation seed.
    Smaller values indicate closer groups in CLIP feature space.}
    \label{tbl:appx_flux_clip_distances}
\end{table}
\begin{table}[t]
    \centering
    \small
    \setlength{\tabcolsep}{14pt}
    \begin{tabular}{lcc}
        \toprule
        Corruption & Setting & Acc. \\
        \midrule
        None & - & 100.00 \\
        \midrule
        Color jittering & Strength: 1.0 & 100.00 \\
        Color jittering & Strength: 2.0 & 99.94 \\
        \midrule
        Gaussian blur & Radius: 1.0 & 100.00 \\
        Gaussian blur & Radius: 2.0 & 100.00 \\
        Gaussian blur & Radius: 4.0 & 100.00 \\
        Gaussian blur & Radius: 8.0 & 100.00 \\
        \midrule
        Gaussian noise & Blending weight: 0.2 & 100.00 \\
        Gaussian noise & Blending weight: 0.4 & 99.91 \\
        Gaussian noise & Blending weight: 0.6 & 99.59 \\
        Gaussian noise & Blending weight: 0.8 & 97.99 \\
        \midrule
        Low resolution & Resolution: 64$\times$64 & 100.00 \\
        Low resolution & Resolution: 32$\times$32 & 100.00 \\
        Low resolution & Resolution: 16$\times$16 & 99.94 \\
        Low resolution & Resolution: 8$\times$8 & 99.21 \\
        Low resolution & Resolution: 4$\times$4 & 95.60 \\
        Low resolution & Resolution: 2$\times$2 & 84.74 \\
        \midrule
        Patch shuffle & Patch size: 32$\times$32 & 99.43 \\
        Patch shuffle & Patch size: 16$\times$16 & 97.02 \\
        Patch shuffle & Patch size: 8$\times$8 & 92.49 \\
        \midrule
        Reconstruction & SD-VAE & 100.00 \\
        Reconstruction & Qwen-VAE & 100.00 \\
        \bottomrule
    \end{tabular}
    \caption{\textbf{Full corruption and transformation results.} For each setting, a ResNet-50 is trained and evaluated on the same transformed ImageNet-derived distribution. Accuracy remains high under moderate corruptions, stays well above chance under the most severe settings, and is essentially unaffected by patch shuffling with large patches or by \acs{vae} reconstruction.}
    \label{tbl:appx_corruption_full}
\end{table}
\begin{table}[t]
    \centering
    \small
    \setlength{\tabcolsep}{13.5pt}
    \begin{tabular}{llc}
        \toprule
        Task & Classifier & Acc. \\
        \midrule
        Four-way & ResNet-50 & \(69.58\pm3.66\) \\
        Four-way & Full MLP (225D) & \(97.92\pm0.36\) \\
        Four-way & Texture MLP (38D) & \(98.33\pm0.36\) \\
        \midrule
        Eight-way & Texture MLP (38D) & \(97.29\pm0.18\) \\
        \bottomrule
    \end{tabular}
    \caption{\textbf{Source attribution for API-generated images.}
    Four-way classification uses the four API services, and eight-way classification adds the four main open-source generators.
    Each source contributes 157 training images and 40 validation images from disjoint categories.
    Values are validation accuracies in percent, reported as means \(\pm\) sample standard deviations across training seeds 42, 43, and 44 at the final epoch.
    The ResNet-50 is randomly initialized.}
    \label{tbl:appx_closed_source}
\end{table}
\begin{table}[t]
    \centering
    \small
    \setlength{\tabcolsep}{8.5pt}
    \begin{tabular}{lcc}
        \toprule
        Model pair & $n$ (no-hint/hint) & Effect \\
        \midrule
        \acs{flux}--\acs{hidream} & 17/11 & 9.57 \\
        \acs{flux}--\acs{hunyuan} & 10/9 & 12.11 \\
        \acs{flux}--\acs{qwenimage} & 10/10 & 5.00 \\
        \acs{hidream}--\acs{hunyuan} & 11/10 & 10.00 \\
        \acs{hidream}--\acs{qwenimage} & 11/10 & 10.55 \\
        \acs{hunyuan}--\acs{qwenimage} & 10/11 & $-0.82$ \\
        \bottomrule
    \end{tabular}
    \caption{\textbf{Pair-specific hint effects.} Effects are hint minus no-hint mean accuracy in percentage points. Five of the six pair-specific effects are positive.}
    \label{tbl:appx_hint_by_pair}
\end{table}
\begin{table*}[t]
    \centering
    \footnotesize
    \setlength{\tabcolsep}{12.5pt}
    \begin{tabular}{@{}lrp{0.72\textwidth}@{}}
        \toprule
        Category & Dim. & Descriptor composition \\
        \midrule
        Tone & 32 & Four channel-average intensity moments; 16-bin HSV-value histogram; seven luminance quantiles; two clipping ratios; luminance entropy; two local RMS-contrast statistics. \\
        Color & 60 & Twelve RGB moments; 16-bin hue and saturation histograms; eight CIELAB $a^\ast,b^\ast$ moments; three RGB correlations; colorfulness; mean sine and cosine of hue; two Lab-chroma statistics. \\
        Texture & 38 & Eight \acs{glcm} statistics; 10-bin $(P,R)=(8,1)$ and 18-bin $(P,R)=(16,2)$ uniform-\acs{lbp} histograms; two local-entropy statistics. \\
        Edge and sharpness & 20 & Three Canny densities; three gradient-magnitude statistics; four \acs{hog} summaries; 9-bin magnitude-weighted Sobel-orientation histogram; Laplacian variance. \\
        Frequency & 21 & \acs{fft} high/low-energy ratio; four radial energy bands; 12 mean Gabor amplitudes; spectral slope, centroid, spread, and flatness. \\
        Spatial organization & 40 & Nine Otsu-mask morphology statistics; 18 cell-wise luminance statistics and nine edge densities on a $3\times3$ grid; two reflection errors; two luminance-centroid coordinates. \\
        Residual and noise & 14 & Four high-pass residual moments; three absolute-residual quantiles; three one-pixel autocorrelations; four radial energy bands. \\
        \midrule
        \textbf{Total} & \textbf{225} & \textbf{Seven mutually exclusive feature categories.} \\
        \bottomrule
    \end{tabular}
    \caption{\textbf{Composition of the 225-dimensional handcrafted feature bank.} Features are extracted from each native $1024\times1024$ image. ``Dim.'' denotes the number of scalar descriptors assigned to each mutually exclusive category. All descriptors are computed per image before standardization with ImageNet-derived training statistics.}
    \label{tbl:appx_feature_bank}
\end{table*}

\subsection{Real ImageNet Images}\label{sec:appx_real_images}

For the real-synthetic similarity analysis (\cref{tbl:real_synthetic_similarity}), we sample 10,000 real ImageNet images covering all 1,000 classes, with one to eleven images per class. Each real image passes through the same pipelines as the generated images: the handcrafted feature extractor of \cref{sec:appx_low_level}, SAM~3.1 queried with the class label, MoGe-2, DiffusionRenderer, and Qwen3-VL-8B-Instruct as the \ac{vlm} captioner under both prompts of \cref{sec:appx_captioning}. Each representation is then fed to the four-way classifier trained on that representation. For the segmentation-mask representation, the mask-area filter of \cref{sec:appx_mid_level_statistics} retains 7,653 images from 982 classes. The sampled real images are originally square. We convert them to RGB and resize them to \(1024\times1024\) using Lanczos interpolation, saving them as JPEG with quality 95 and no chroma subsampling. Handcrafted features are extracted directly from these prepared images without further resizing or cropping.

\section{Additional Results on Source Classification}\label{sec:appx_source_classification}

\subsection{Confusion Matrices}\label{sec:appx_confusion_main}

\Cref{fig:appx_confusion_main} reports the row-normalized confusion matrices of the four classifiers on the held-out ImageNet-derived test set. The two end-to-end \acp{cnn} produce strongly diagonal matrices: ResNet-50 reaches 100.00\% accuracy and the two-layer \ac{cnn} 98.81\%, with the latter's residual errors concentrated in \ac{flux} images misclassified as \ac{hidream}. Linear classifiers on frozen DINOv2 and CLIP features remain well above the 25\% chance level but show broader confusion: DINOv2 most often confuses \ac{flux} with \ac{hidream}, whereas CLIP assigns images from several sources to \ac{flux}. Source identity is thus most readily recovered by end-to-end visual classifiers, while frozen semantic features retain a weaker source signal.

\subsection{Robustness to \ac{flux} Generation Settings}
\label{sec:appx_flux_settings}

We evaluate \ac{flux} under five settings of inference steps and distilled guidance.
Each setting uses the same 200 ImageNet-derived evaluation prompts and generation seeds 42, 66, 222, and 999, yielding 800 images at \(1024\times1024\) resolution.
Each \ac{flux} batch is evaluated alongside the other three generators using the frozen four-way ResNet-50 from the main experiments with \ac{hidream}.
\Cref{tbl:appx_flux_settings} shows that \ac{flux} remains reliably identifiable across the tested settings.

We additionally compare image distributions using unit-normalized frozen CLIP ViT-B/32 features.
Centroid distance is the Euclidean distance between group means.
Gaussian-kernel \(\mathrm{MMD}^{2}\), with bandwidth \(\sigma=1\), also captures distributional differences beyond the mean.
Using the newly generated \ac{flux} \((50,4)\) batch as the reference, \cref{tbl:appx_flux_clip_distances} shows smaller differences for the four setting variants than for the three other generators.

\subsection{Full Corruption Sweeps and Additional Transformations}\label{sec:appx_corruptions}

\Cref{tbl:appx_corruption_full} reports all settings of the four corruptions in the main text, of which \cref{tbl:corruption_robustness} shows a subset. Each transformation is applied to both the training and the held-out images, and a separate ResNet-50 is trained and evaluated on each transformed distribution. Gaussian blur is implemented with the \texttt{GaussianBlur} filter of Pillow, whose only parameter is the blur radius in pixels; the kernel size follows from the radius. Examples of corrupted images are shown in \cref{fig:appx_corruption_examples}.

We additionally consider two transformations. Patch shuffle divides each image into non-overlapping $p\times p$ patches and randomly permutes their spatial locations; classifiers are trained and evaluated on shuffled images. Accuracy decreases from 99.43\% with $32\times32$ patches to 92.49\% with $8\times8$ patches. Local image statistics thus remain highly predictive after global spatial organization is disrupted, while increasingly local shuffling removes useful structure. Reconstruction encodes and decodes every image with the same off-the-shelf \ac{vae} before training and evaluation. ResNet-50 reaches 100.00\% with both the Stable Diffusion \ac{vae}~\citep{rombach2022high} and the \ac{qwenimage} \ac{vae}~\citep{qwen2025qwenimage2512}. The source signal therefore survives a shared learned reconstruction bottleneck and cannot be explained solely by fragile final-stage decoder artifacts.

\subsection{Source Attribution for Closed-Source Models}
\label{sec:appx_closed_source}

We extend the experiment to GPT-Image-2.5 Sunburst and Qwen-Image-3.0-Pro through their official APIs, and Gemini Nano Banana 2.1 and Grok Imagine Image 2.0 through Poe.
We reuse the original class prompts and collect native \(1024\times1024\) outputs, with GPT quality set to high, Grok quality set to medium, and Qwen prompt expansion and visible watermark disabled in the request.
We retain 197 categories with successful outputs from all four services, using one image per source and category.
A fixed random partition assigns 157 categories to training and 40 to validation, shared across sources.
Classifiers are trained from scratch for 100 epochs, and feature normalization is fitted on training images.
For the eight-way extension, we add one matched generation-seed-42 image per category from each of the four main open-source generators, including \ac{hidream}.
\Cref{tbl:appx_closed_source} shows strong texture-based attribution in both the four-way and eight-way tasks, extending the evidence to native API outputs under these service configurations.

\subsection{Statistical Analysis of the Human Study}\label{sec:appx_human_stats}

\paragraph{Effect of the hint.}
To test the effect of the hint while respecting the blocked design, we treat each participant's mean accuracy over the ten main trials as one observation and use a two-sided stratified permutation test. Under the null hypothesis of no hint effect, we independently permute the hint and no-hint labels within each model-pair-by-mapping block, preserving the observed condition counts in every block. Across 50,000 permutations, the observed difference of 9.03 percentage points (71.64\% with the hint versus 62.61\% without) has $p<0.001$. A stratified nonparametric bootstrap that resamples participants within each model-pair-by-mapping-by-condition cell (50,000 resamples) gives a 95\% confidence interval of $[4.76, 13.25]$ percentage points.

As a trial-level robustness check, we fit a Bayesian mixed-effects logistic regression to the 1,300 binary main-trial responses, with hint condition, model pair, mapping, and centered trial index as fixed effects and random intercepts for participant and query image. With weakly informative $\mathcal{N}(0,5^2)$ priors on the fixed effects and $\mathcal{N}(0,1)$ priors on the log random-effect standard deviations, the estimated odds ratio of the hint is 1.51, with a 95\% credible interval of $[1.25, 1.82]$.

\paragraph{Consistency across pairs and exclusion rules.}
\Cref{tbl:appx_hint_by_pair} reports the hint effect within each model pair. Five of the six effects are positive, so the overall effect is not driven by a single pair. The result is also robust to the exclusion rule: excluding only the participant who fails both attention checks retains 149 participants (73 hint and 76 no-hint) and yields a difference of 8.45 percentage points (stratified permutation $p<0.001$), and including all 150 submissions (73 hint and 77 no-hint) yields 8.61 percentage points ($p<0.001$).

\begin{figure*}[t]
    \centering
    \begin{subfigure}[t]{0.3\textwidth}
        \centering
        \includegraphics[width=\linewidth]{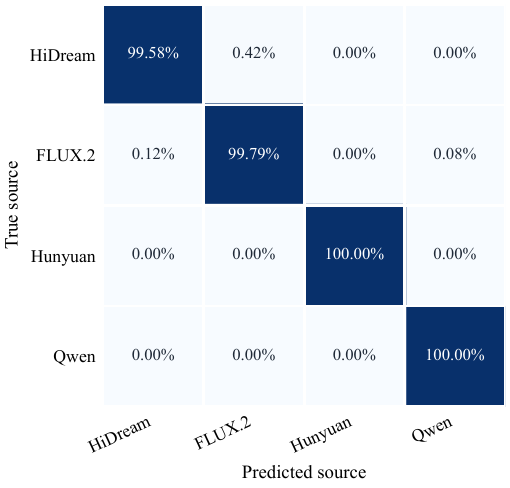}
        \caption{All features.}
    \end{subfigure}
    \hfill
    \begin{subfigure}[t]{0.3\textwidth}
        \centering
        \includegraphics[width=\linewidth]{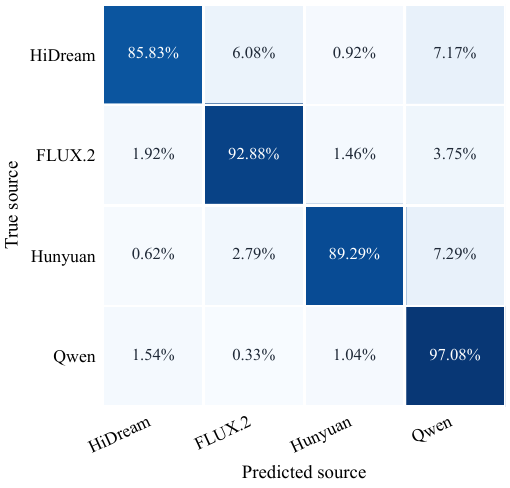}
        \caption{Tone only.}
    \end{subfigure}
    \hfill
    \begin{subfigure}[t]{0.3\textwidth}
        \centering
        \includegraphics[width=\linewidth]{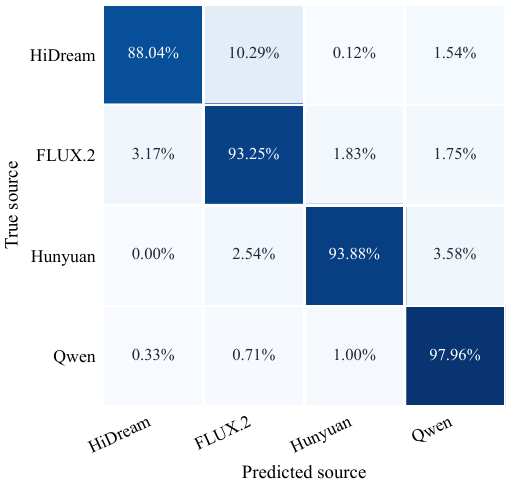}
        \caption{Color only.}
    \end{subfigure}

    \begin{subfigure}[t]{0.3\textwidth}
        \centering
        \includegraphics[width=\linewidth]{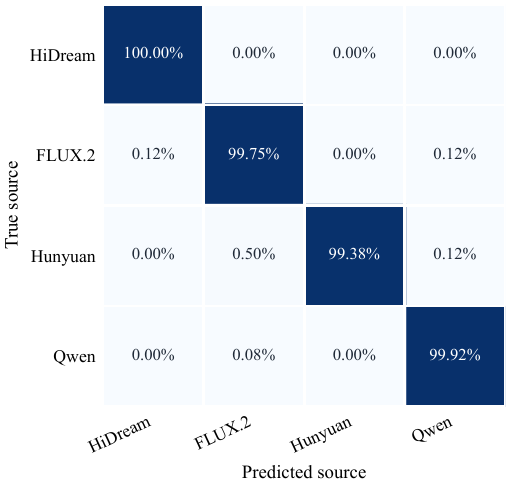}
        \caption{Texture only.}
    \end{subfigure}
    \hfill
    \begin{subfigure}[t]{0.3\textwidth}
        \centering
        \includegraphics[width=\linewidth]{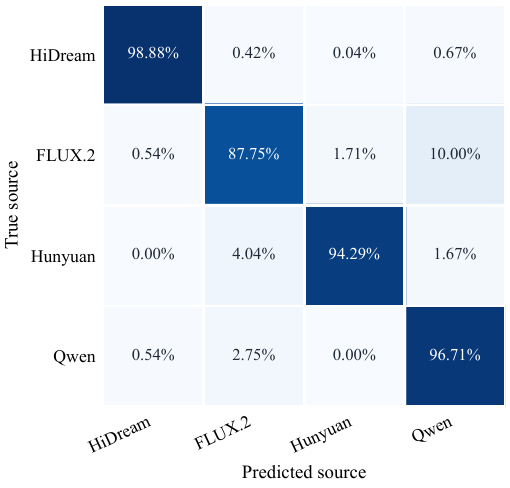}
        \caption{Edge and sharpness only.}
    \end{subfigure}
    \hfill
    \begin{subfigure}[t]{0.3\textwidth}
        \centering
        \includegraphics[width=\linewidth]{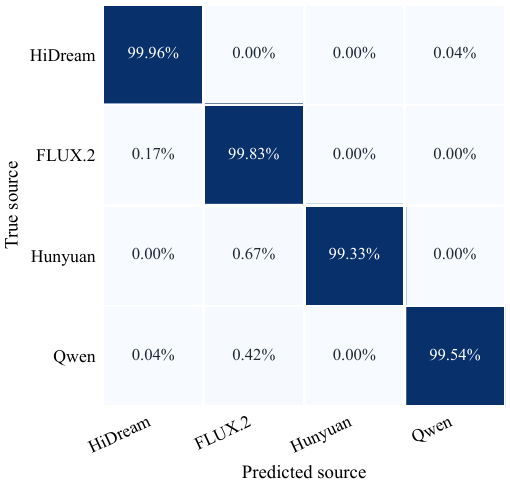}
        \caption{Frequency only.}
    \end{subfigure}

    \begin{subfigure}[t]{0.3\textwidth}
        \centering
        \includegraphics[width=\linewidth]{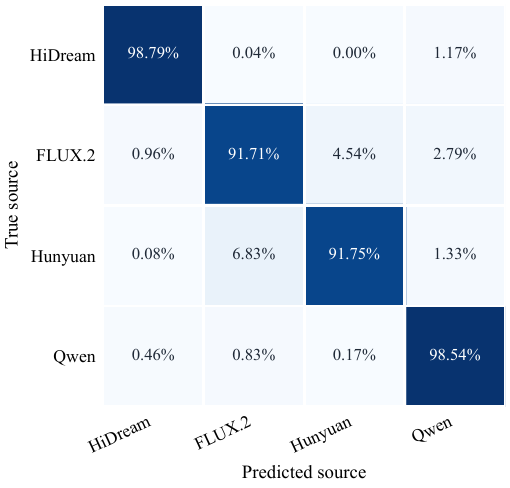}
        \caption{Spatial organization only.}
    \end{subfigure}
    \qquad
    \begin{subfigure}[t]{0.3\textwidth}
        \centering
        \includegraphics[width=\linewidth]{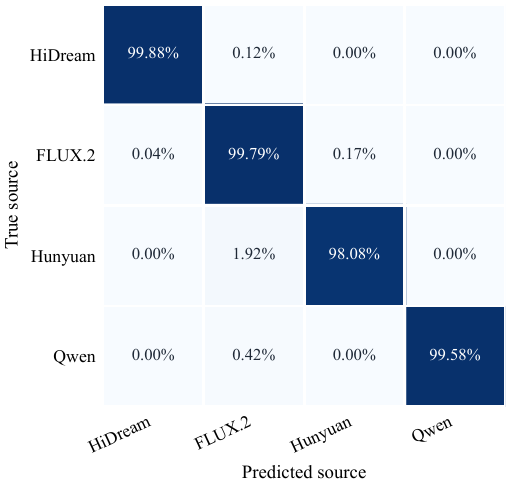}
        \caption{Residual and noise only.}
    \end{subfigure}
    \caption{\textbf{Source-classification confusion matrices for the full handcrafted feature bank and for each feature category alone.} Rows denote true source models and columns denote predicted sources. Each cell reports the row-normalized percentage, aggregated over three training seeds on the held-out ImageNet-derived test set.}
    \label{fig:appx_confusion_low_level}
\end{figure*}

\begin{figure*}[t]
    \centering
    \begin{subfigure}[t]{0.8\textwidth}
        \centering
        \includegraphics[width=\linewidth]{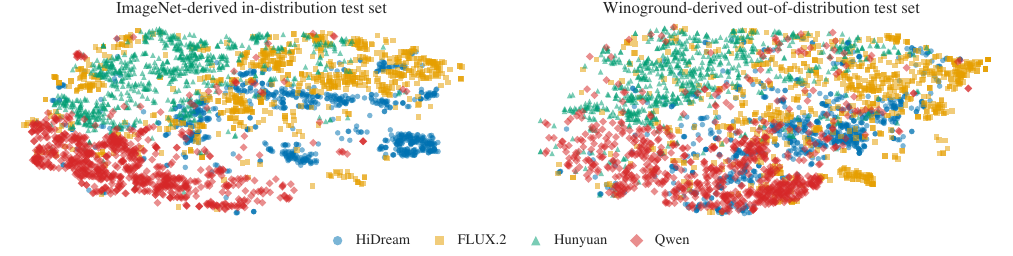}
        \caption{Tone.}
    \end{subfigure}

    \begin{subfigure}[t]{0.8\textwidth}
        \centering
        \includegraphics[width=\linewidth]{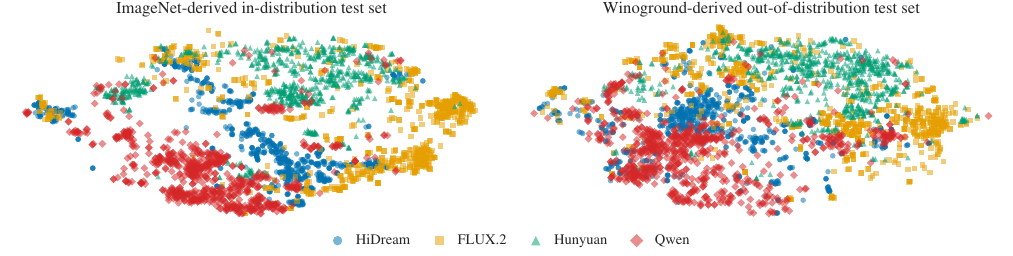}
        \caption{Color.}
    \end{subfigure}
    \caption{\textbf{t-SNE visualization of each handcrafted feature category.} A joint embedding is fitted to balanced \acs{id} and \acs{ood} subsets and displayed separately per domain with shared coordinates. t-SNE is applied directly to the standardized features without PCA preprocessing. Continued in \cref{fig:appx_tsne_low_level_cont}.}
    \label{fig:appx_tsne_low_level}
\end{figure*}
\begin{figure*}[tp]
    \centering
    \ContinuedFloat
    \begin{subfigure}[t]{0.8\textwidth}
        \centering
        \includegraphics[width=\linewidth]{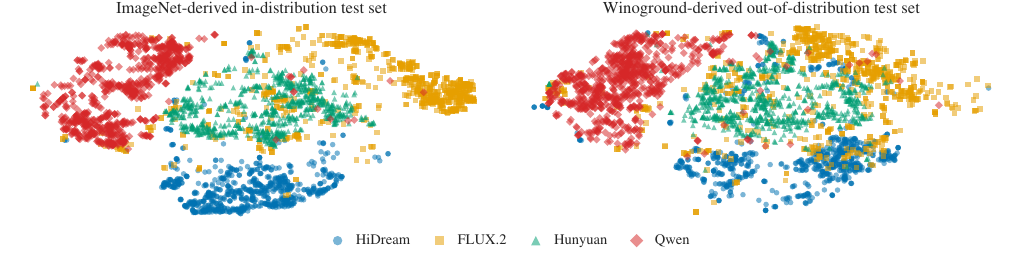}
        \caption{Texture.}
    \end{subfigure}

    \begin{subfigure}[t]{0.8\textwidth}
        \centering
        \includegraphics[width=\linewidth]{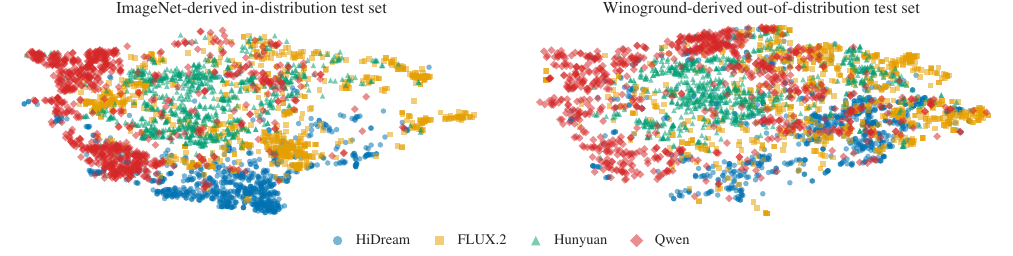}
        \caption{Edge and sharpness.}
    \end{subfigure}

    \begin{subfigure}[t]{0.8\textwidth}
        \centering
        \includegraphics[width=\linewidth]{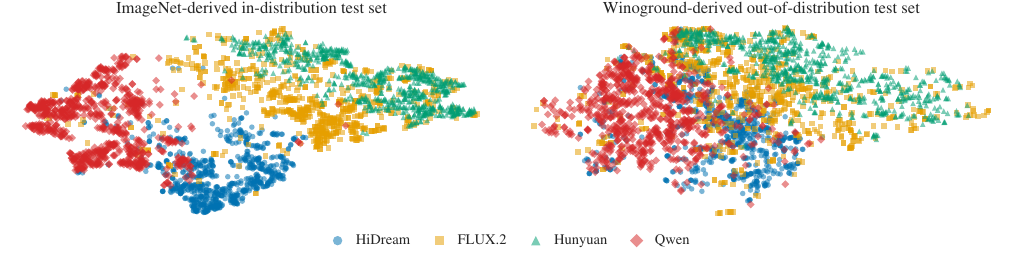}
        \caption{Frequency.}
    \end{subfigure}
    
    \begin{subfigure}[t]{0.8\textwidth}
        \centering
        \includegraphics[width=\linewidth]{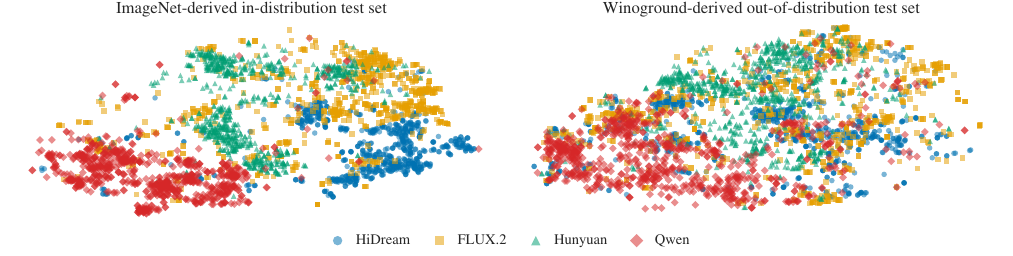}
        \caption{Spatial organization.}
    \end{subfigure}

    \begin{subfigure}[t]{0.8\textwidth}
        \centering
        \includegraphics[width=\linewidth]{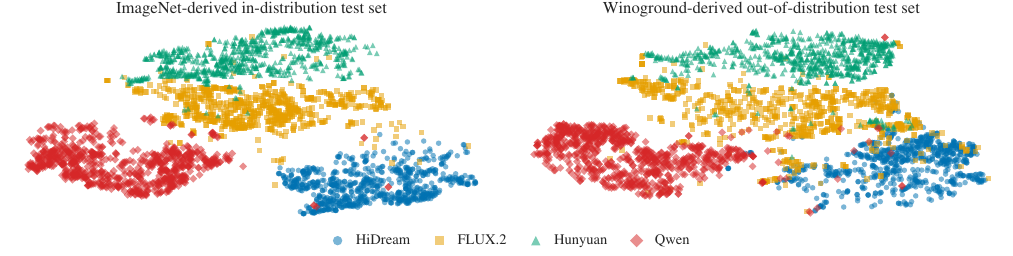}
        \caption{Residual and noise.}
    \end{subfigure}
    \caption{\textbf{t-SNE visualization of each handcrafted feature category (continued).} Texture is the only category whose embedding separates all four source models in both domains.}
    \label{fig:appx_tsne_low_level_cont}
\end{figure*}
\begin{figure*}[t]
    \centering
    \begin{subfigure}[t]{0.5\textwidth}
        \centering
        \includegraphics[width=\linewidth]{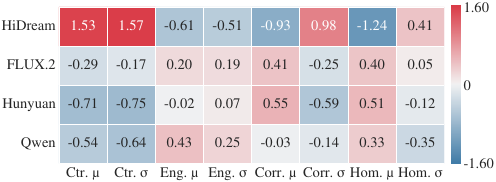}
        \caption{\acs{glcm} statistics on the ImageNet-derived set.}
    \end{subfigure}%
    \begin{subfigure}[t]{0.5\textwidth}
        \centering
        \includegraphics[width=\linewidth]{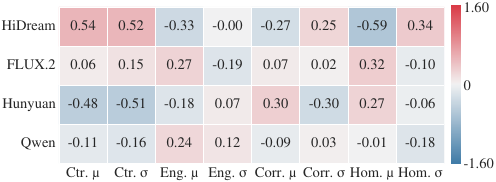}
        \caption{\acs{glcm} statistics on the Winoground-derived set.}
    \end{subfigure}
    \caption{\textbf{Prompt-centered \acs{glcm} statistics.} Cells report source-wise mean prompt-centered z-scores for the mean and standard deviation of \acs{glcm} contrast, energy, correlation, and homogeneity. Source-specific signed profiles remain visible on both prompt distributions.}
    \label{fig:appx_glcm}
\end{figure*}
\begin{figure*}[t]
    \centering
    \begin{subfigure}[t]{0.5\textwidth}
        \centering
        \includegraphics[width=\linewidth]{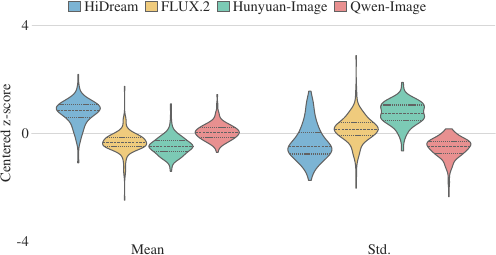}
        \caption{Local entropy on the ImageNet-derived set.}
    \end{subfigure}%
    \begin{subfigure}[t]{0.5\textwidth}
        \centering
        \includegraphics[width=\linewidth]{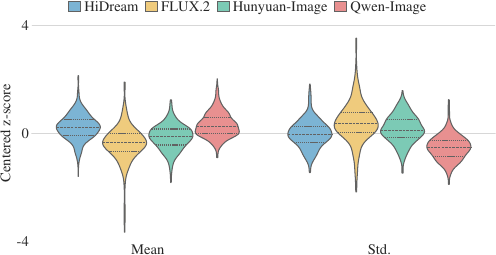}
        \caption{Local entropy on the Winoground-derived set.}
    \end{subfigure}
    \caption{\textbf{Prompt-centered local-entropy distributions.} Violin plots show the distributions of the per-image local-entropy mean and standard deviation after matched-prompt centering. \ac{hidream}’s entropy mean is above the four-source average in both domains. Its entropy standard deviation is below this average on ImageNet and close to it on Winoground.}
    \label{fig:appx_local_entropy}
\end{figure*}

\section{Low-Level Visual Cues}\label{sec:appx_low_level}

\subsection{Feature Bank Composition}\label{sec:appx_feature_bank}

We extract a 225-dimensional handcrafted feature vector from every image directly from the original $1024\times1024$ PNG, without resizing or cropping. The bank comprises seven disjoint categories: tone (32 dimensions), color (60), texture (38), edge and sharpness (20), frequency (21), spatial organization (40), and residual and noise (14). A fixed registry assigns every descriptor to exactly one category (\cref{tbl:appx_feature_bank}).

\subsection{Feature Definitions}\label{sec:appx_feature_definitions}

\paragraph{Shared definitions.}
For normalized color channels $R,G,B\in[0,1]$, we define the channel-average intensity $I$ and the luminance $Y$ as
\begin{equation}
    \begin{aligned}
        I &= (R+G+B)/3, \\
        Y &= 0.2126R+0.7152G+0.0722B.
    \end{aligned}
\end{equation}
For values $x_1,\ldots,x_N$, let $\mu=N^{-1}\sum_i x_i$ and $m_k=N^{-1}\sum_i(x_i-\mu)^k$. We report the mean $\mu$, variance $m_2$, skewness $\gamma_1$, and excess kurtosis $\gamma_2$, where
\begin{equation}
    \gamma_1 = \frac{m_3}{(m_2+\epsilon)^{3/2}+\epsilon},
    \qquad
    \gamma_2 = \frac{m_4}{m_2^2+\epsilon}-3,
\end{equation}
with $\epsilon=10^{-12}$. Histograms are normalized to unit mass.

\paragraph{Tone and color.}
The 32 tone descriptors comprise four moments of $I$, a 16-bin HSV-value histogram, seven luminance quantiles ($1,5,25,50,75,95,99\%$), dark and bright clipping ratios at $1/255$ and $254/255$, 256-bin luminance entropy~\citep{shannon1948mathematical}, and the mean and standard deviation of $16\times16$ local RMS contrast. The 60 color descriptors comprise moments of each RGB channel, 16-bin hue and saturation histograms, moments of CIELAB $a^\ast$ and $b^\ast$~\citep{luo2023cielab}, the three pairwise RGB correlations, Hasler--S\"usstrunk colorfulness~\citep{hasler2003measuring}, the mean sine and cosine of hue, and the mean and standard deviation of Lab chroma.

\paragraph{Texture and edges.}
The 38 texture descriptors include \acp{glcm}~\citep{haralick1973textural}, uniform \acp{lbp}~\citep{ojala2002multiresolution}, and local entropy. \acp{glcm} use 32 quantized levels of $I$, distances $\{1,2\}$, and orientations $\{0,\pi/4,\pi/2,3\pi/4\}$; we retain the mean and standard deviation of contrast, energy, correlation, and homogeneity. Uniform \acp{lbp} contribute 10 bins at $(P,R)=(8,1)$ on $I$ and 18 bins at $(16,2)$ on quantized $Y$, followed by the mean and standard deviation of radius-5 local entropy on $Y$. The 20 edge and sharpness descriptors comprise Canny densities~\citep{canny1986computational} at $\sigma\in\{1,2,4\}$, the gradient-magnitude mean, standard deviation, and maximum, four \ac{hog} summaries~\citep{dalal2005histograms} (mean, standard deviation, maximum, and nonzero fraction), a magnitude-weighted 9-bin Sobel-orientation histogram, and Laplacian variance. The $\sigma=1$ Canny, gradient, and \ac{hog} features use $I$, while the remaining descriptors use $Y$; \ac{hog} uses 9 orientations, $16\times16$ cells, $2\times2$ blocks, and L2-Hys normalization.

\paragraph{Frequency.}
The 21 frequency descriptors comprise an \ac{fft} high/low-energy ratio on $I$ with cutoff $0.1\min(H,W)$, normalized energy in four equal-width radial bands of the same spectrum, mean Gabor amplitudes~\citep{gabor1946theory} on $I$ for four orientations and three frequencies $\{0.1,0.2,0.3\}$, and four power-spectrum summaries of $Y$. For the latter, let $P=|\mathcal{F}(Y)|^2$, let $r\in[0,1]$ denote normalized radial frequency, and let $Z=\sum P$ over non-DC coefficients:
\begin{equation}
    \begin{aligned}
        c &= \textstyle\sum rP/Z, \\
        s &= \left[\textstyle\sum(r-c)^2P/Z\right]^{1/2}, \\
        f &= \exp(\mathbb{E}[\log(P+\epsilon)])/(\mathbb{E}[P]+\epsilon).
    \end{aligned}
\end{equation}
These quantities are the spectral centroid, spread, and flatness. The spectral slope is fitted by linear regression of log mean power on log radial frequency over 32 radial bins, excluding the innermost bin.

\paragraph{Spatial organization and residual.}
The 40 spatial-organization descriptors combine morphology and coarse layout. An Otsu mask~\citep{otsu1979threshold} of $I$ provides mask occupancy, connected-component count, the mean and standard deviation of normalized component area, solidity, and circularity, and mean extent. A $3\times3$ grid contributes the luminance mean, luminance standard deviation, and Canny-$\sigma=1$ density of every cell. We additionally compute mean absolute left--right and top--bottom reflection errors of $Y$ and the luminance-weighted centroid in normalized image coordinates. The 14 residual and noise descriptors use $R=Y-\mathcal{G}_{\sigma=1}\ast Y$: four moments, the 50th, 90th, and 99th percentiles of $|R|$, one-pixel horizontal, vertical, and diagonal autocorrelations, and normalized energy in four radial \ac{fft} bands.

\paragraph{Implementation.}
We implement the extractor with NumPy~\citep{harris2020array}, OpenCV~\citep{bradski2008learning}, and scikit-image~\citep{vanderwalt2014scikitimage}, and the classifier in PyTorch~\citep{paszke2019pytorch}. Raw feature vectors are stored as \texttt{float32}. All 22,400 generated images are verified to be $1024\times1024$, and all extracted values are finite.

\subsection{Handcrafted-Feature Classifier}\label{sec:appx_low_level_classifier}

Features follow the data split of \cref{sec:appx_classifiers}. Each feature is standardized with ImageNet-derived training statistics,
\begin{equation}
    \widetilde{x}_{ij} = (x_{ij}-\mu_j^{\mathrm{train}})/\sigma_j^{\mathrm{train}},
\end{equation}
where $\sigma_j^{\mathrm{train}}<10^{-6}$ is set to one; this affects only \texttt{residual\_moment\_mean}. The \ac{mlp} is $d\!\rightarrow\!256\!\rightarrow\!128\!\rightarrow\!4$, where $d$ is the input dimension, with ReLU and dropout~\citep{srivastava2014dropout} of 0.1 after both hidden layers. We train with cross-entropy, Adam~\citep{adam} at $10^{-4}$, batch size 16, and 100 epochs, retaining the checkpoint with the highest validation accuracy. Hyperparameters are fixed across configurations. Reported accuracies are means over seeds 42, 43, and 44.

\subsection{Feature-Category Analysis}\label{sec:appx_category_analysis}

We evaluate each feature category from two complementary perspectives. For category sufficiency, we train the same \ac{mlp} using only the descriptors in that category. For grouped permutation importance~\citep{breiman2001random,gregorutti2015grouped}, we evaluate the trained full-bank classifier after applying a common random row permutation to all descriptors in one category while leaving the other categories unchanged; this preserves dependencies within the permuted category but breaks its association with the image and its source label. Permutations are performed separately for each dataset and training seed. Both analyses use the same train-only normalization, data splits, architecture, and checkpoint-selection procedure described above.

\subsection{Confusion Matrices and t-SNE Embeddings}\label{sec:appx_low_level_viz}

\Cref{fig:appx_confusion_low_level} reports the row-normalized confusion matrices for the complete feature bank and each feature category alone. The complete bank achieves 99.84\% ID test accuracy, while texture, frequency, and residual and noise features alone reach 99.76\%, 99.67\%, and 99.33\%, respectively. Tone, color, edge and sharpness, and spatial organization reach 91.27--95.20\%. Class-wise performance varies with the feature category. \ac{hidream} has the highest recall for edge and sharpness (98.88\%) and spatial organization (98.79\%), computed by pooling confusion counts across the three training seeds. Multiple feature categories therefore independently retain strong source information.

We visualize each feature category with t-SNE~\citep{van2008visualizing}. For every category, we standardize the corresponding feature dimensions and apply t-SNE directly, without a preceding PCA projection. A single joint embedding is fitted to balanced ImageNet-derived \ac{id} and Winoground-derived \ac{ood} subsets, and the two domains are displayed separately with the same embedding coordinates, which makes changes in source separation across prompt distributions directly visible. As shown in \cref{fig:appx_tsne_low_level,fig:appx_tsne_low_level_cont}, texture features form a clear four-source organization in both domains. Frequency and residual-and-noise features retain partial source structure on the ImageNet-derived set, but several sources overlap substantially, and the overlap grows on Winoground. Tone, color, edge and sharpness, and spatial-organization embeddings overlap extensively and do not form four distinct clusters.

\begin{figure*}[t]
    \centering
    \begin{subfigure}[t]{0.5\textwidth}
        \centering
        \includegraphics[width=\linewidth]{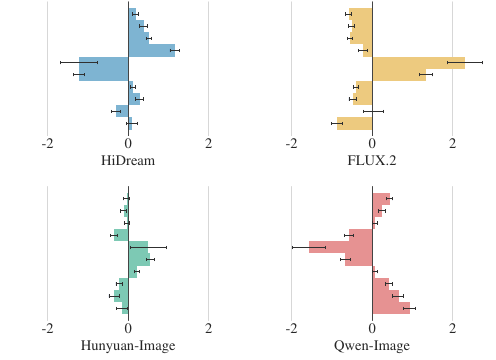}
        \caption{ImageNet-derived set, $(P,R)=(8,1)$.}
    \end{subfigure}%
    \begin{subfigure}[t]{0.5\textwidth}
        \centering
        \includegraphics[width=\linewidth]{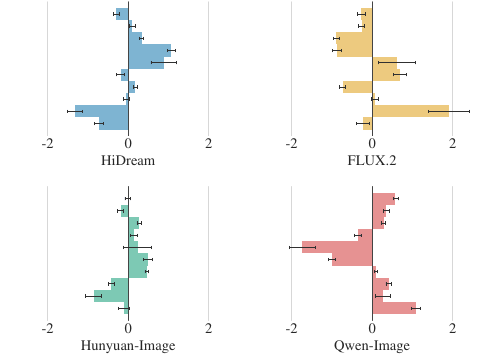}
        \caption{Winoground-derived set, $(P,R)=(8,1)$.}
    \end{subfigure}

    \begin{subfigure}[t]{0.5\textwidth}
        \centering
        \includegraphics[width=\linewidth]{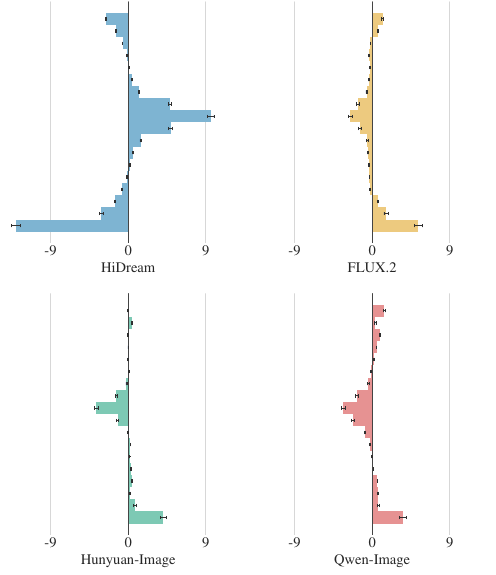}
        \caption{ImageNet-derived set, $(P,R)=(16,2)$.}
    \end{subfigure}%
    \begin{subfigure}[t]{0.5\textwidth}
        \centering
        \includegraphics[width=\linewidth]{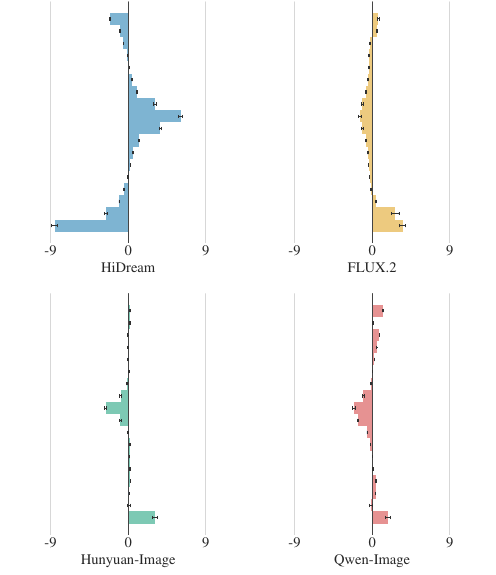}
        \caption{Winoground-derived set, $(P,R)=(16,2)$.}
    \end{subfigure}
    \caption{\textbf{Prompt-centered uniform-\acs{lbp} profiles.} Within each model panel, horizontal bars are ordered by uniform-\acs{lbp} bin index from top to bottom: bins 0--9 for $(P,R)=(8,1)$ and bins 0--17 for $(P,R)=(16,2)$, with the final bin collecting non-uniform patterns. For every seed-prompt group and bin, we subtract the mean bin probability across the four source models and average the centered values over the four seeds of each prompt. Bar lengths report the mean prompt-centered bin-probability difference across prompts in percentage points, with zero denoting the matched four-source mean.}
    \label{fig:appx_lbp}
\end{figure*}

\subsection{Texture Statistics}\label{sec:appx_texture_stats}

We examine the constituent statistics of the texture category in greater detail. Before aggregation, each feature is centered across the four source models within every matched seed-prompt group, and the four seeds of each prompt are then averaged for each source. This removes the component shared by images generated from the same prompt and emphasizes systematic source-conditioned offsets rather than prompt-content differences.

As shown in \cref{fig:appx_glcm}, the \ac{glcm} statistics reveal distinct co-occurrence profiles for the four source models. \ac{hidream} and \ac{hunyuan} tend to occupy opposite directions on several contrast- and energy-related statistics, while \ac{flux} and \ac{qwenimage} exhibit different correlation and homogeneity offsets. The magnitudes change between ImageNet and Winoground, but the signed source-dependent profiles remain visible in both domains, so the signal captured by \ac{glcm} features is not reducible to a single notion of contrast or smoothness.

\cref{fig:appx_local_entropy} shows complementary differences in local entropy. \ac{hidream}’s local-entropy mean is above the four-source average on both ImageNet and Winoground, with average prompt-centered z-scores of +0.765 and +0.227, respectively. Its local-entropy standard deviation is below this average on ImageNet (-0.343) and close to it on Winoground (-0.022). \ac{qwenimage} also has positive entropy-mean offsets and negative entropy-standard-deviation offsets on both prompt distributions.

The uniform \ac{lbp} histograms in \cref{fig:appx_lbp} provide a more localized view of texture. At both $(P,R)=(8,1)$ and $(P,R)=(16,2)$, the four source models assign systematically different probability mass to specific pattern bins, with the largest deviations concentrated in a small subset of bins rather than spread over the histogram. Several prominent offsets, including the distinctive \ac{hidream} and \ac{flux} profiles, persist under the Winoground prompt shift even though their amplitudes change. Together, the \ac{glcm}, local-entropy, and \ac{lbp} results show that texture-based source classification combines several forms of local structure rather than relying on one scalar texture statistic.

\begin{figure*}[t]
    \centering
    \begin{subfigure}[t]{0.4\textwidth}
        \centering
        \includegraphics[width=\linewidth]{fig_resnet50_confusion_matrix}
        \caption{Original image.}
    \end{subfigure}
    \qquad
    \begin{subfigure}[t]{0.4\textwidth}
        \centering
        \includegraphics[width=\linewidth]{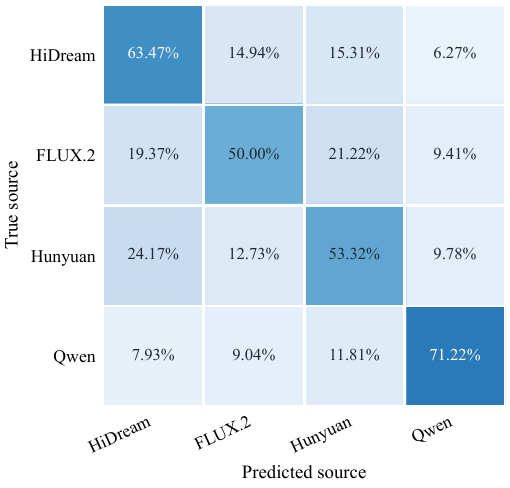}
        \caption{Segmentation mask.}
    \end{subfigure}

    \begin{subfigure}[t]{0.4\textwidth}
        \centering
        \includegraphics[width=\linewidth]{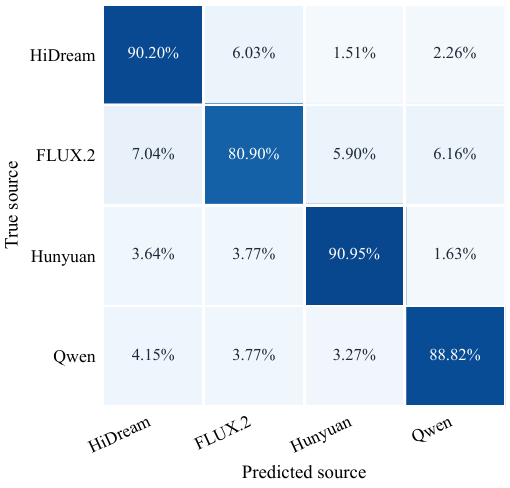}
        \caption{Depth map.}
    \end{subfigure}
    \qquad
    \begin{subfigure}[t]{0.4\textwidth}
        \centering
        \includegraphics[width=\linewidth]{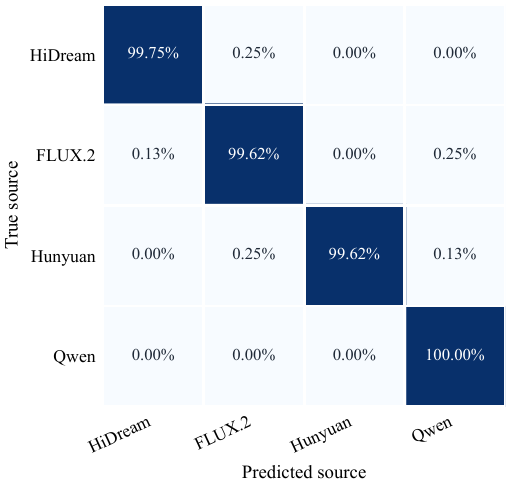}
        \caption{Albedo map.}
    \end{subfigure}
    \caption{\textbf{Source-classification confusion matrices for the mid-level visual representations.} All four classifiers are ResNet-50. Rows denote true source models and columns denote predicted sources. Each cell reports the row-normalized percentage.}
    \label{fig:appx_confusion_mid_level}
\end{figure*}

\section{Mid-Level Visual Representations}\label{sec:appx_mid_level}

\subsection{Representation Extraction}\label{sec:appx_mid_level_extraction}

We extract foreground masks, metric depth, and albedo from every generated image. For foreground masks, we use text-guided segmentation with SAM~3.1~\citep{carion2026sam}. On ImageNet-derived images, we query SAM~3.1 once with the corresponding ImageNet class label and take the pixelwise union of all returned instance masks. On Winoground-derived images, we query SAM~3.1 separately with the two extracted object names and take the pixelwise union of the masks returned by both queries. In both cases, the foreground mask is defined by the alpha channel of the resulting cutout.

We apply the off-the-shelf MoGe-2~\citep{wang2025moge} model to each image and retain its per-pixel metric-depth prediction. The depth representation given to the source classifier, however, is the rendered relative-depth image rather than the raw metric-depth array. For each image $i$, we convert every valid metric depth $z_i(p)>0$ to disparity $u_i(p)=1/z_i(p)$ and compute the $0.001$ and $0.99$ disparity quantiles, $q_{0.001}^{(i)}$ and $q_{0.99}^{(i)}$, over all valid pixels. The relative depth is
\begin{equation}
    r_i(p) = \operatorname{clip}\!\left(1-\frac{u_i(p)-q_{0.001}^{(i)}}{q_{0.99}^{(i)}-q_{0.001}^{(i)}}, 0, 1\right).
\end{equation}
We map $r_i$ to RGB with the MoGe-2 depth colormap and use the rendered image as the classifier input. The classification representation is thus normalized independently for every generated image and does not encode absolute metric scale; the raw metric-depth values are used only for the metric-depth analysis in \cref{sec:appx_mid_level_statistics}.

We additionally apply the off-the-shelf DiffusionRenderer~\citep{liang2025diffusion} inverse-rendering model and retain its estimated base-color pass as the albedo map.

\subsection{Layout, Depth, and Albedo Statistics}\label{sec:appx_mid_level_statistics}

\paragraph{Average object layouts.}
We first remove masks whose foreground occupies less than 5\% or more than 95\% of the image. To preserve the matched comparison, if any source model fails this criterion for a given prompt and seed, we remove the corresponding images from all four sources, leaving 2,612 matched images per source model. For each retained $1024\times1024$ mask $M_i$, we approximate its signed distance field in pixel units as
\begin{equation}
    d_i(p) = \operatorname{EDT}\!\left(M_i\right)(p) - \operatorname{EDT}\!\left(1-M_i\right)(p),
\end{equation}
where $\operatorname{EDT}$ is the Euclidean distance transform, so that $d_i(p)$ is positive inside the foreground and negative outside it. For source model $s$, the average layout plotted in \cref{fig:average_layout} is
\begin{equation}
    \bar d_s(p) = \frac{1}{N_s}\sum_{i=1}^{N_s} d_i(p),
\end{equation}
computed directly in the original image coordinates without spatial alignment. We display $\bar d_s$ together with its zero-level contour; the reported mean-layout area is the fraction of pixels satisfying $\bar d_s(p)\geq0$.

\paragraph{Metric depth of generated objects.}
For image $i$, we define the object region as the intersection of the SAM foreground and the pixels with finite MoGe-2 depth,
\begin{equation}
    \Omega_i = \left\{p : M_i(p)=1\right\} \cap \left\{p : z_i(p)\ \text{is finite}\right\},
\end{equation}
where $z_i(p)$ is the predicted metric depth in meters. The near- and far-depth summaries plotted in \cref{fig:geometry_violin_plot} are the 5th and 95th percentiles of $\{z_i(p) : p\in\Omega_i\}$, shown on a logarithmic depth axis.

\paragraph{Fine-scale variation ratio.}
We compute this quantity from the DiffusionRenderer albedo estimate inside the retained SAM foreground. Seven matched groups are excluded because at least one source has insufficient valid pixel pairs at a sampled distance. We remove all four source images in each affected group, leaving 2,605 images per source for this analysis. To reduce boundary artifacts, we erode each foreground by 5 Euclidean pixels and denote the remaining region by $\Omega_i$. Let
\begin{equation}
    L_i = \sqrt{(h_i-1)^2+(w_i-1)^2}
\end{equation}
be the diagonal of the axis-aligned bounding box of $\Omega_i$. For each normalized distance
\begin{equation}
    r \in \{0.005, 0.01, 0.02, 0.04, 0.08, 0.16, 0.24, 0.32, 0.40\},
\end{equation}
we sample 1,024 pixel pairs $(p,q)$ within $\Omega_i$ whose Euclidean separation is approximately $rL_i$. After converting the base-color values to CIELAB, we define the raw albedo variogram as
\begin{equation}
    D_i(r) = \underset{\substack{p,q\in\Omega_i \\ \lVert p-q\rVert_2/L_i\approx r}}{\operatorname{median}} \Delta E_{00}(p,q),
\end{equation}
where $\Delta E_{00}$ is the CIEDE2000 color difference. We fit a nondecreasing curve $\widetilde D_i(r)$ to these values with weighted isotonic regression, using the number of valid pairs as the weight. The fine-scale variation ratio plotted in \cref{fig:appearance_violin_plot} is
\begin{equation}
    R_{\mathrm{fine},i} = \frac{\widetilde D_i(0.005)}{\widetilde D_i(0.40)},
\end{equation}
which measures how much of an object's long-range albedo difference is already present at a separation equal to 0.5\% of the object scale.

\subsection{Confusion Matrices}\label{sec:appx_mid_level_confusion}

\Cref{fig:appx_confusion_mid_level} compares the confusion matrices obtained by ResNet-50 from RGB images and the three estimated mid-level representations. RGB classification reaches 100.00\% ID evaluation accuracy. Albedo achieves 99.75\%, with eight misclassified images among 3,200, and depth achieves 87.72\%. Segmentation masks reach 59.50\% on the 2,168 retained ID images, with \ac{qwenimage} having the highest class-wise recall. These results demonstrate source information in object layout, geometry, and estimated surface appearance.

\subsection{Classification Across Architectures}\label{sec:appx_mid_level_classifiers}

\begin{table}[t]
    \centering
    \small
    \setlength{\tabcolsep}{2pt}
    \begin{tabular}{llcc}
        \toprule
        Representation & Classifier & Acc. (\acs{id}) & Acc. (\acs{ood}) \\
        \midrule
        \multirow{4}{*}{Segmentation mask} & ResNet-50 & 59.50 & 46.73 \\
        & Two-layer \acs{cnn} & 56.78 & 42.29 \\
        & DINOv2 ViT-B/14 & 50.46 & 36.17 \\
        & CLIP ViT-B/32 & 44.56 & 32.46 \\
        \midrule
        \multirow{4}{*}{Depth map} & ResNet-50 & 87.72 & 70.50 \\
        & Two-layer \acs{cnn} & 76.66 & 56.67 \\
        & DINOv2 ViT-B/14 & 69.22 & 56.22 \\
        & CLIP ViT-B/32 & 60.96 & 45.95 \\
        \midrule
        \multirow{4}{*}{Albedo map} & ResNet-50 & 99.75 & 97.73 \\
        & Two-layer \acs{cnn} & 95.41 & 88.58 \\
        & DINOv2 ViT-B/14 & 86.81 & 80.41 \\
        & CLIP ViT-B/32 & 81.66 & 72.08 \\
        \bottomrule
    \end{tabular}
    \caption{\textbf{Source classification from mid-level visual representations across classifier architectures.} \acs{id} denotes evaluation on held-out ImageNet-derived prompts and \acs{ood} denotes evaluation on Winoground-derived prompts without target-domain training. All values are accuracies in percent; chance is 25\%.}
    \label{tbl:appx_mid_level_classifiers}
\end{table}

\Cref{tbl:appx_mid_level_classifiers} reports results for all combinations of the three representations and the four classifier architectures. The ordering of the representations is consistent across every classifier and both prompt distributions: albedo is most predictive, depth intermediate, and segmentation masks least predictive. ResNet-50 gives the strongest result for each representation, but the two-layer \ac{cnn} and the linear classifiers on frozen DINOv2 and CLIP features also remain above the 25\% chance level. Accuracy decreases on Winoground-derived prompts, while the ordering across representations is preserved. Albedo remains highly discriminative, reaching 97.73\% \ac{ood} accuracy with ResNet-50 and 72.08\% with a linear classifier on frozen CLIP features.

\section{High-Level Semantic Realization}\label{sec:appx_high_level}

\subsection{Captioning Setup}\label{sec:appx_captioning}

We use Qwen3-VL-8B-Instruct~\citep{bai2025qwen3}, InternVL3.5-8B~\citep{wang2025internvl3}, and GLM-4.1V-9B-Thinking~\citep{hong2025glm} as captioners. Each captioner processes all 16,000 ImageNet-derived and 6,400 Winoground-derived images independently under each of the two prompts below, yielding 44,800 captions per captioner. The \ac{vlm} input contains only the RGB image and the instruction; source labels, filenames, paths, and the original generation prompt are not provided. We use bfloat16 inference, greedy decoding without sampling, and a maximum of 192 new tokens. For the GLM thinking checkpoint, we suppress chain-of-thought generation and retain only the final answer text. Example captions are shown in \cref{tbl:appx_caption_examples}.

\promptbox{Style-description prompt}{%
    \texttt{Describe the image's overall visual style and atmosphere in one or two sentences. Focus on how the image looks and feels as a whole rather than listing the objects or actions it contains. Use only directly observable qualities.}%
}

\promptbox{Background-description prompt}{%
    \texttt{Describe only the visible background of this image in one or two factual sentences. Focus on the setting and on what appears behind or around the main subject, including the background's overall visual appearance. Do not describe or identify the foreground subject. If the background is absent, blurred, or unclear, state that briefly instead of inferring details.}%
}

\begin{figure*}[tp]
    \centering
    \begin{subfigure}[t]{0.4\textwidth}
        \centering
        \includegraphics[width=\linewidth]{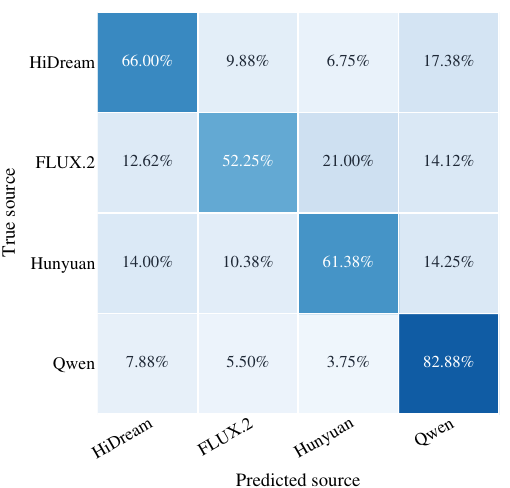}
        \caption{Qwen3-VL-8B, style.}
    \end{subfigure}
    \qquad
    \begin{subfigure}[t]{0.4\textwidth}
        \centering
        \includegraphics[width=\linewidth]{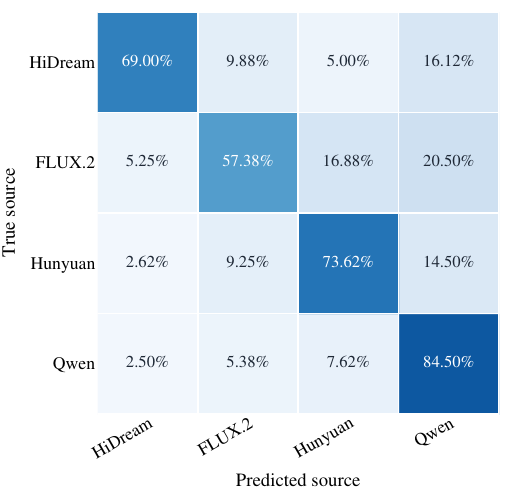}
        \caption{Qwen3-VL-8B, background.}
    \end{subfigure}

    \begin{subfigure}[t]{0.4\textwidth}
        \centering
        \includegraphics[width=\linewidth]{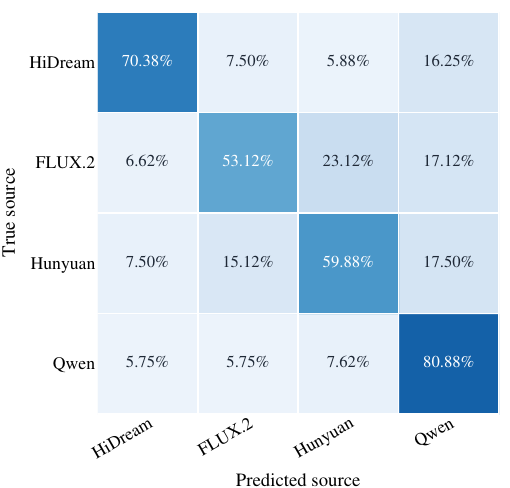}
        \caption{InternVL3.5-8B, style.}
    \end{subfigure}
    \qquad
    \begin{subfigure}[t]{0.4\textwidth}
        \centering
        \includegraphics[width=\linewidth]{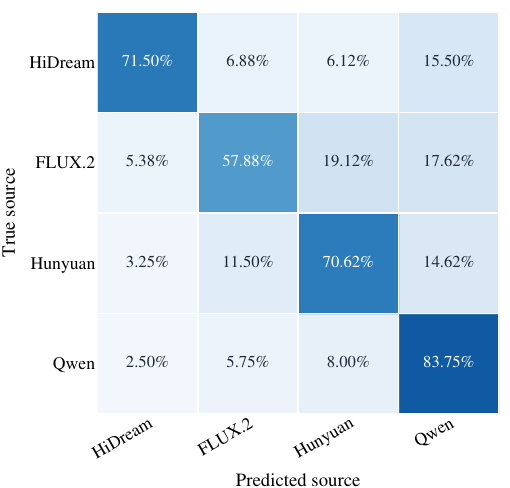}
        \caption{InternVL3.5-8B, background.}
    \end{subfigure}

    \begin{subfigure}[t]{0.4\textwidth}
        \centering
        \includegraphics[width=\linewidth]{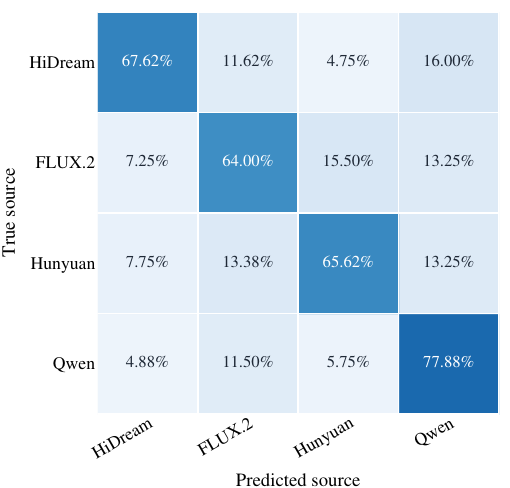}
        \caption{GLM-4.1V-9B, style.}
    \end{subfigure}
    \qquad
    \begin{subfigure}[t]{0.4\textwidth}
        \centering
        \includegraphics[width=\linewidth]{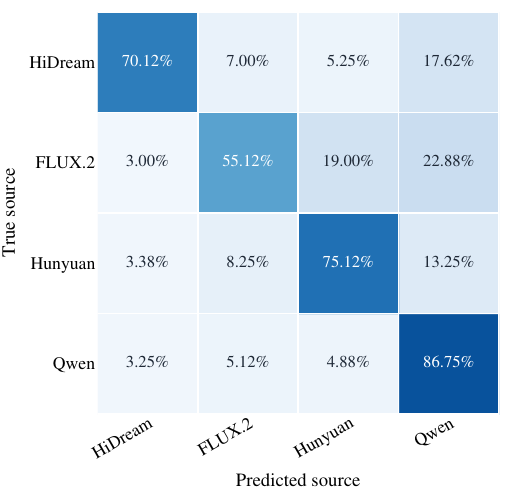}
        \caption{GLM-4.1V-9B, background.}
    \end{subfigure}
    \caption{\textbf{Caption-based source-classification confusion matrices for style and background descriptions from three \acs{vlm} captioners.} Each classifier uses combined word- and character-level \acs{tfidf} features and is evaluated on the held-out ImageNet-derived test set. Rows denote true source models and columns denote predicted sources. Each cell reports the row-normalized percentage.}
    \label{fig:appx_confusion_high_level}
\end{figure*}

\subsection{Caption Classification}\label{sec:appx_caption_classifier}

The source-classification task has the same four classes and the same image-level splits as \cref{sec:appx_classifiers}; captions never determine the split.

\paragraph{Text preprocessing.}
Style and background captions are analyzed separately for every captioner; we do not concatenate captions across prompts or captioners. We remove leading and trailing whitespace and, for GLM, its answer-formatting markers. We otherwise preserve the generated text: no manual stop-word removal, stemming, lemmatization, or punctuation filtering is applied. Lowercasing is performed by the \ac{tfidf} vectorizers, with Unicode accent normalization for the word stream.

\paragraph{Word- and character-level features.}
We concatenate two sparse \ac{tfidf} representations~\citep{salton1988term} implemented with scikit-learn~\citep{pedregosa2011scikit}. The word branch uses unigrams and bigrams, while the character branch uses within-word character 3- to 5-grams. Both branches use sublinear term frequency, smoothed inverse document frequency, $L_2$ normalization, a minimum document frequency of two, and at most 100,000 features; the word branch additionally applies Unicode accent stripping. For each captioner-prompt condition, both vocabularies and all inverse-document-frequency statistics are fitted only on the ImageNet-derived training captions and then held fixed for validation, held-out \ac{id} testing, and Winoground \ac{ood} testing.

\paragraph{Logistic regression.}
For each of the six captioner-prompt conditions, we fit a separate four-class, $L_2$-regularized multinomial logistic-regression classifier with the L-BFGS solver~\citep{liu1989limited}, an iteration limit of 1,000, and random seed 0, without class reweighting. The training set contains 12,800 captions; the held-out \ac{id}, and \ac{ood} sets contain 3,200, and 6,400 captions, respectively. Accuracy is the primary metric, with chance at 25\%. We select the inverse regularization strength $C\in\{0.1,1,10\}$ by ImageNet-derived validation accuracy, independently for each captioner and prompt, and evaluate the selected classifier unchanged on the held-out \ac{id} and \ac{ood} sets. The selected value is $C=1$ for both Qwen3-VL conditions, both InternVL3.5 conditions, and GLM-4.1V background, and $C=10$ for GLM-4.1V style.

\subsection{Confusion Matrices}\label{sec:appx_high_level_confusion}

\Cref{fig:appx_confusion_high_level} reports the row-normalized confusion matrices of the combined word- and character-level \ac{tfidf} classifiers on the held-out ImageNet-derived test set. All six matrices are diagonally dominant. On the ID test set, style and background descriptions yield accuracies of 65.62\% and 71.12\% for Qwen3-VL, 66.06\% and 70.94\% for InternVL3.5, and 68.78\% and 71.78\% for GLM-4.1V, respectively. Background descriptions yield higher accuracy for every captioner.

Across the six combinations of captioner and description prompt, \ac{qwenimage} has the highest recall (77.88--86.75\%) and \ac{flux} has the lowest (52.25--64.00\%). The largest off-diagonal entry for \ac{flux} is assignment to \ac{hunyuan} in four conditions and to \ac{qwenimage} in two. The dominant confusion direction therefore varies with the captioner and description prompt.

\subsection{Model-Specific Preferences on Style}\label{sec:appx_style_words}

\begin{table}[t]
    \centering
    \small
    \setlength{\tabcolsep}{4pt}
    \begin{tabular}{@{}llll@{}}
        \toprule
        \acs{hidream} & \acs{flux} & \hunyuanstack & \qwenimagestack \\
        \midrule
        soft natural & \textbf{smooth} & hazy & \textbf{dark} \\
        window & minimalist & dusty & \textbf{green} \\
        brick & overcast & \textbf{dreamy} & moody \\
        natural light & golden & bokeh & damp \\
        naturalistic & clean & dust & verdant \\
        \midrule
        background of & image features & warm & \textbf{dark} \\
        wooden & minimalist & \textbf{dreamy} & \textbf{green} \\
        window & simple & sunlight & \textbf{green} background \\
        lush & features & warm soft & against \textbf{dark} \\
        greenery & \textbf{smooth} & gentle & muted \\
        \midrule
        natural and & has warm & \textbf{dreamy} & moody \\
        lush & \textbf{smooth} & dusty & has moody \\
        homey & warm soft & sunlit & \textbf{green} \\
        cozy & clean & warm & intimate \\
        brick & overcast & hazy & \textbf{dark} \\
        \bottomrule
    \end{tabular}
    \caption{\textbf{Model-specific style preferences revealed by \acs{vlm}-generated captions.} We list the top-5 predictive word n-grams from word-level \acs{tfidf} logistic-regression classifiers for style descriptions. The three row blocks correspond, from top to bottom, to Qwen3-VL-8B, InternVL3.5-8B, and GLM-4.1V-9B. Bold words are unigrams that rank among the top terms for the same source model under all three captioners.}
    \label{tbl:appx_style_words}
\end{table}

To interpret the lexical evidence underlying style-based source classification, we fit an auxiliary word-level \ac{tfidf} logistic-regression classifier for each captioner and report in \cref{tbl:appx_style_words} the five terms with the largest positive coefficient for each source model; the corresponding results for background descriptions are in \cref{tbl:high_level_semantics_word}. The table retains both unigrams and bigrams, consistent with the word-level representation used in the classification experiments. Coefficients indicate predictive associations and should not be interpreted causally.

Several associations recur across all three captioners. \ac{flux} is associated with \textit{smooth}, \ac{hunyuan} with \textit{dreamy}, and \ac{qwenimage} with \textit{dark} and \textit{green}. \ac{hidream}'s top-five lists vary across captioners, with terms including \textit{soft natural}, \textit{window}, \textit{brick}, \textit{lush}, \textit{homey}, and \textit{cozy}. These coefficients describe source-associated language patterns, with agreement depending on the model and captioner.

\section{Qualitative Examples}\label{sec:appx_examples}

\begin{figure*}[t]
    \centering
    \begin{subfigure}[t]{\textwidth}
        \centering
        \includegraphics[width=\linewidth]{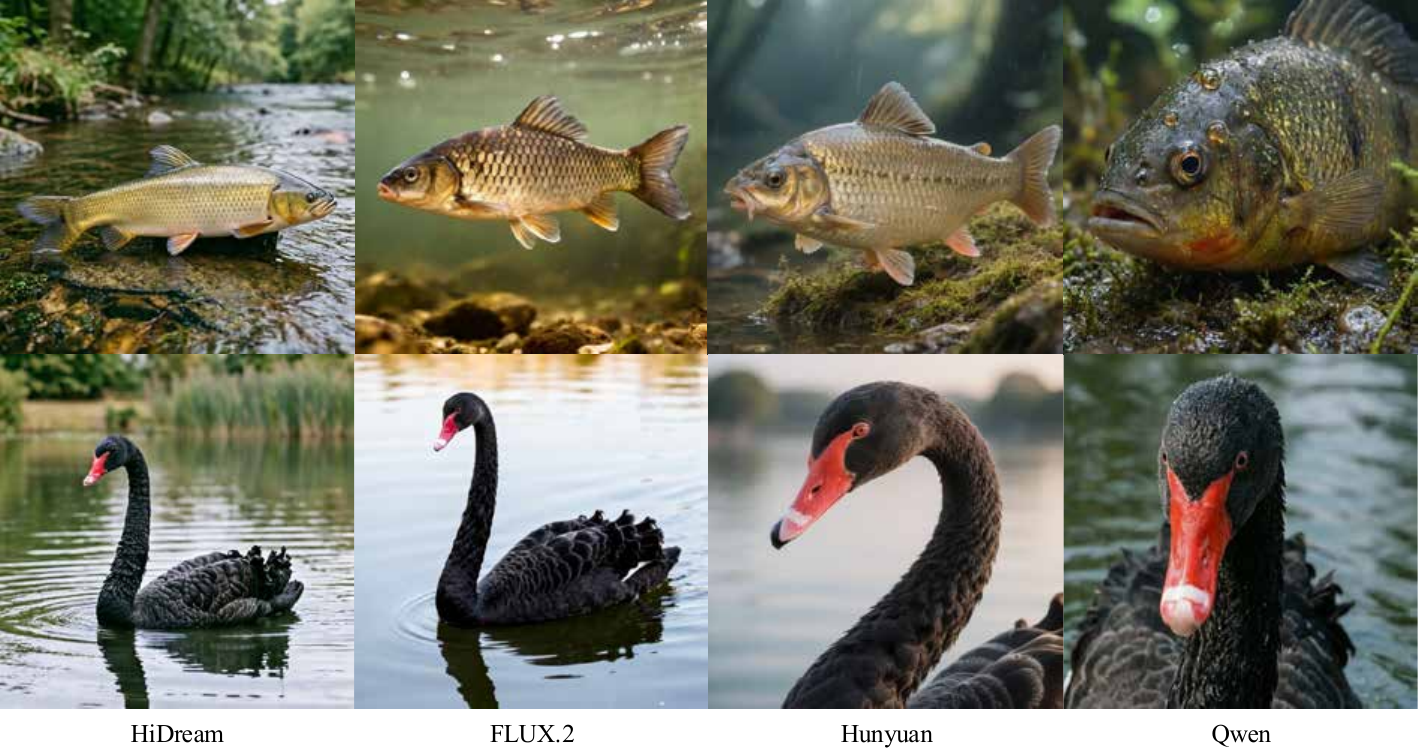}
        \caption{ImageNet-derived prompts. Above: ``a photo of tench, which is also known as Tinca tinca.'' Below: ``a photo of black swan, which is also known as Cygnus atratus.''}
    \end{subfigure}

    \begin{subfigure}[t]{\textwidth}
        \centering
        \includegraphics[width=\linewidth]{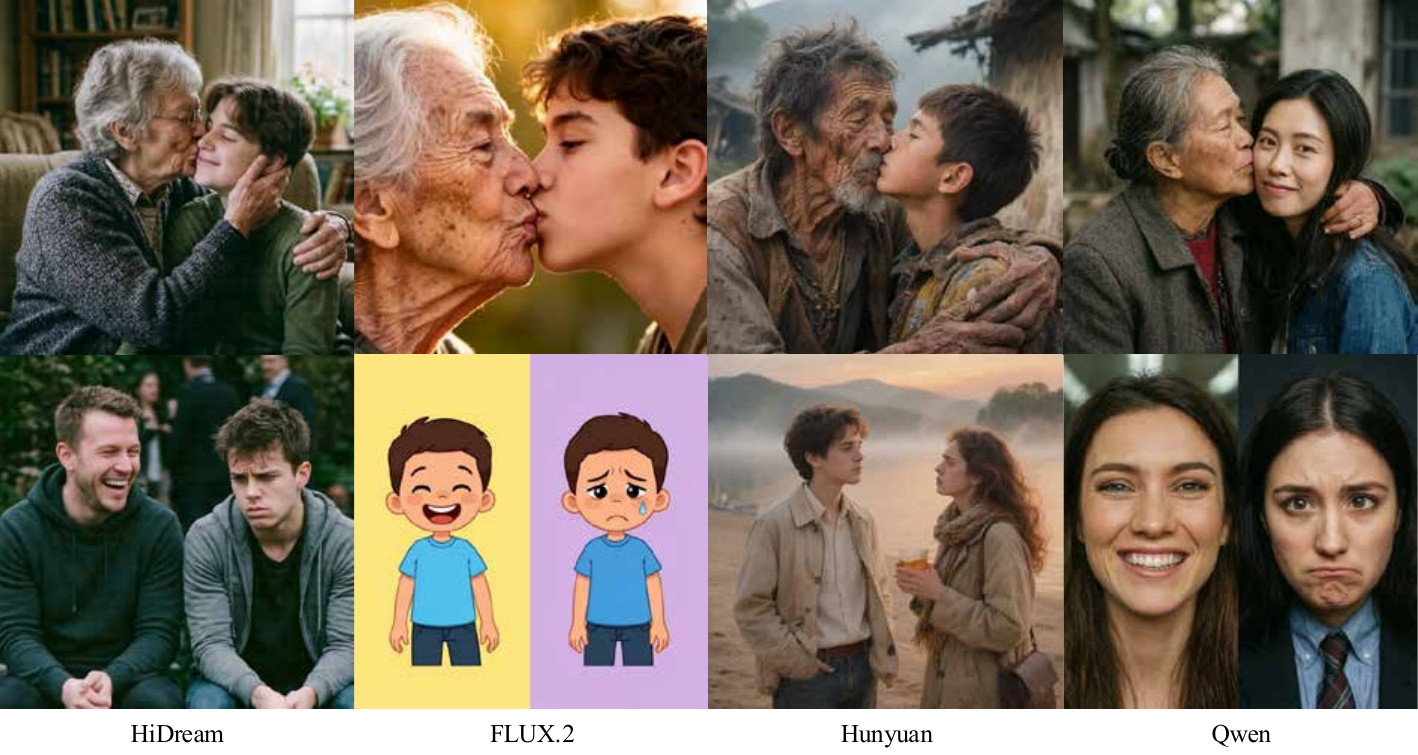}
        \caption{Winoground-derived prompts. Above: ``an old person kisses a young person.'' Below: ``the happy one is on the left and the upset one is on the right.''}
    \end{subfigure}
    \caption{\textbf{Examples of generated images.} Each row shows the four source models' outputs for the same prompt.}
    \label{fig:appx_image_examples}
\end{figure*}

\begin{figure*}[t]
    \centering
    \includegraphics[width=\textwidth]{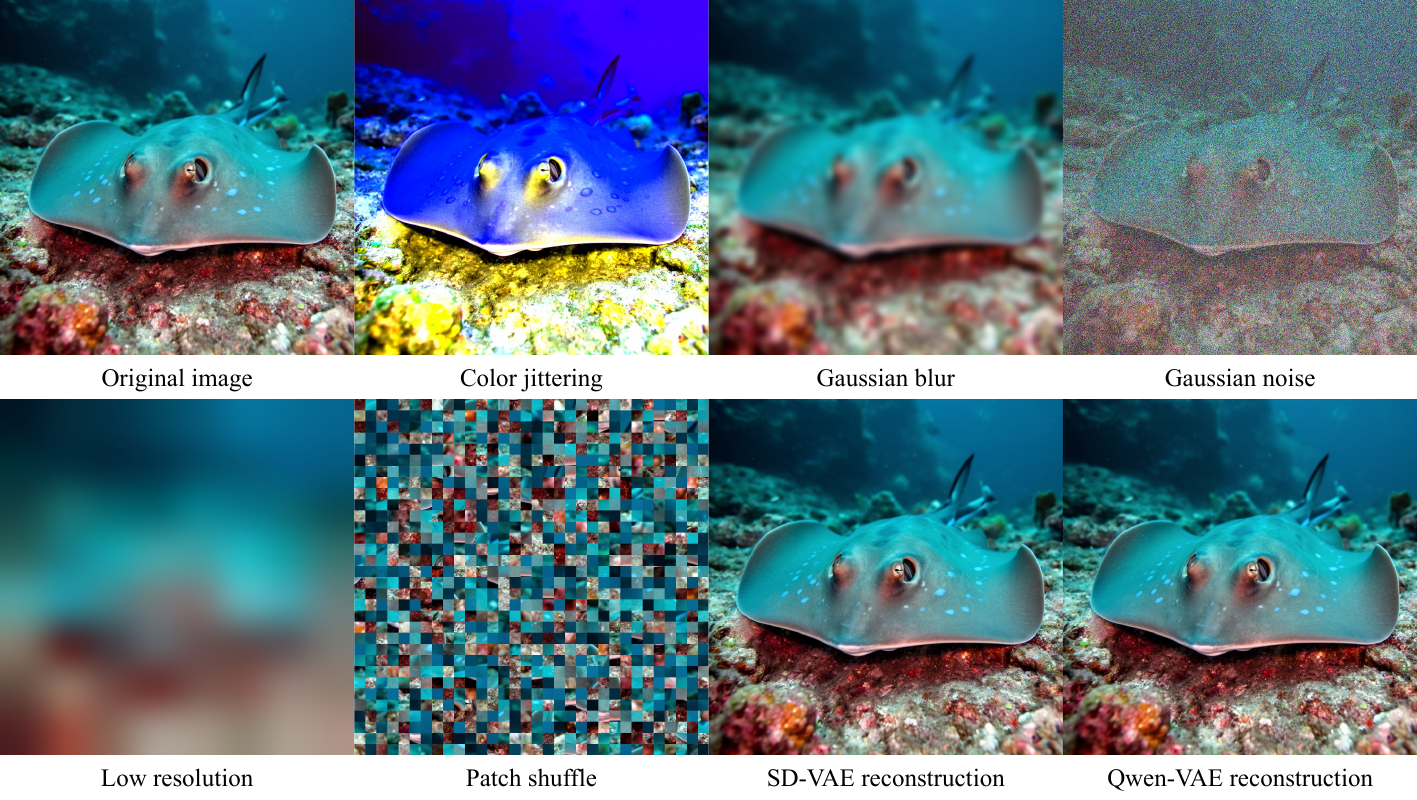}
    \caption{\textbf{Examples of corrupted images.} Each column applies one corruption setting of \cref{tbl:appx_corruption_full} to the same image.}
    \label{fig:appx_corruption_examples}
\end{figure*}

\begin{figure*}[t]
    \centering
    \includegraphics[width=\textwidth]{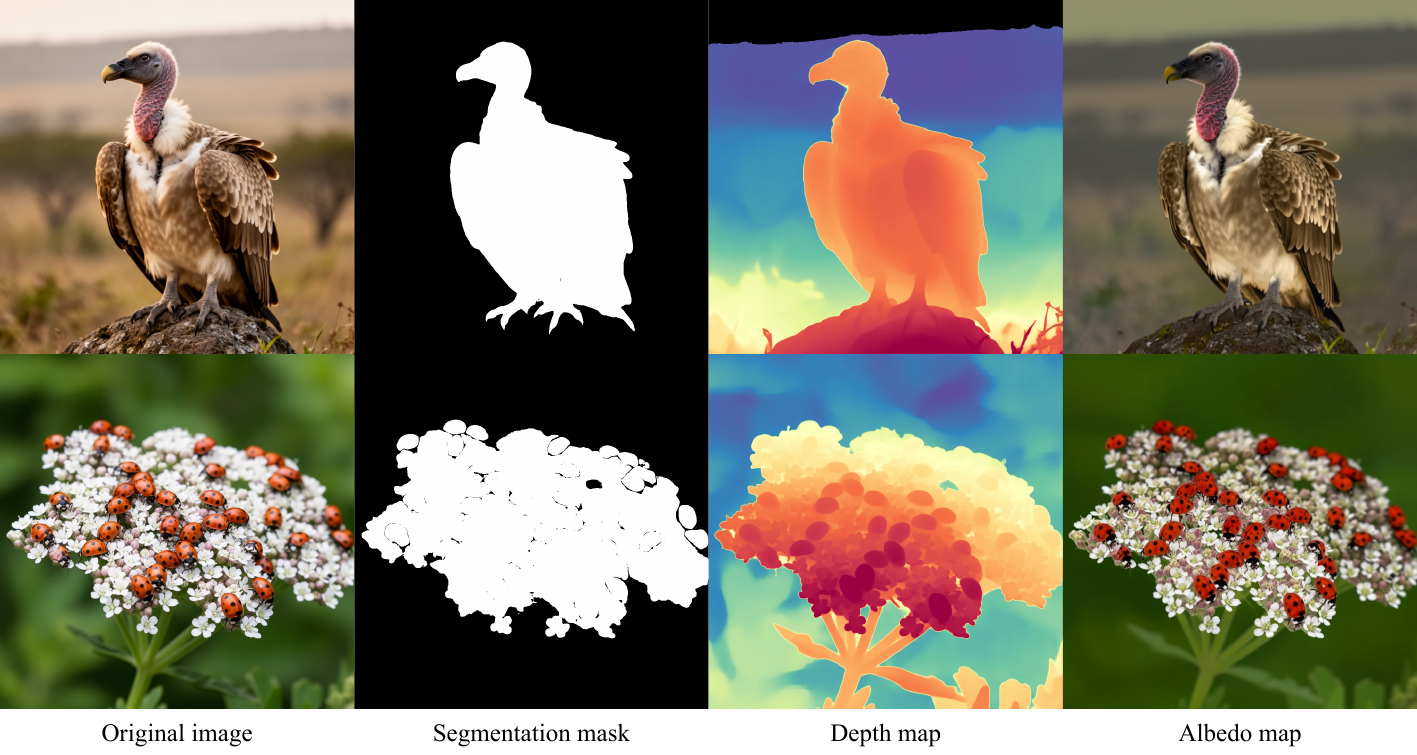}
    \caption{\textbf{Examples of mid-level representations.} Segmentation masks, relative-depth renderings, and albedo maps for images generated from ImageNet-derived prompts (above) and Winoground-derived prompts (below).}
    \label{fig:appx_mid_level_examples}
\end{figure*}

\Cref{fig:appx_image_examples} shows images generated from ImageNet-derived and Winoground-derived prompts, \cref{fig:appx_corruption_examples} shows the corruptions of \cref{sec:appx_corruptions}, \cref{fig:appx_mid_level_examples} visualizes the mid-level representations of \cref{sec:appx_mid_level_extraction}, and \cref{tbl:appx_caption_examples} lists captions produced under the two prompts of \cref{sec:appx_captioning}.

\begin{table*}[t]
    \centering
    \setlength{\tabcolsep}{3pt}
    \begin{tabular}{cl}
        \toprule
        Image & Caption \\
        \midrule
        \raisebox{-0.5\height}{\includegraphics[width=0.215\textwidth]{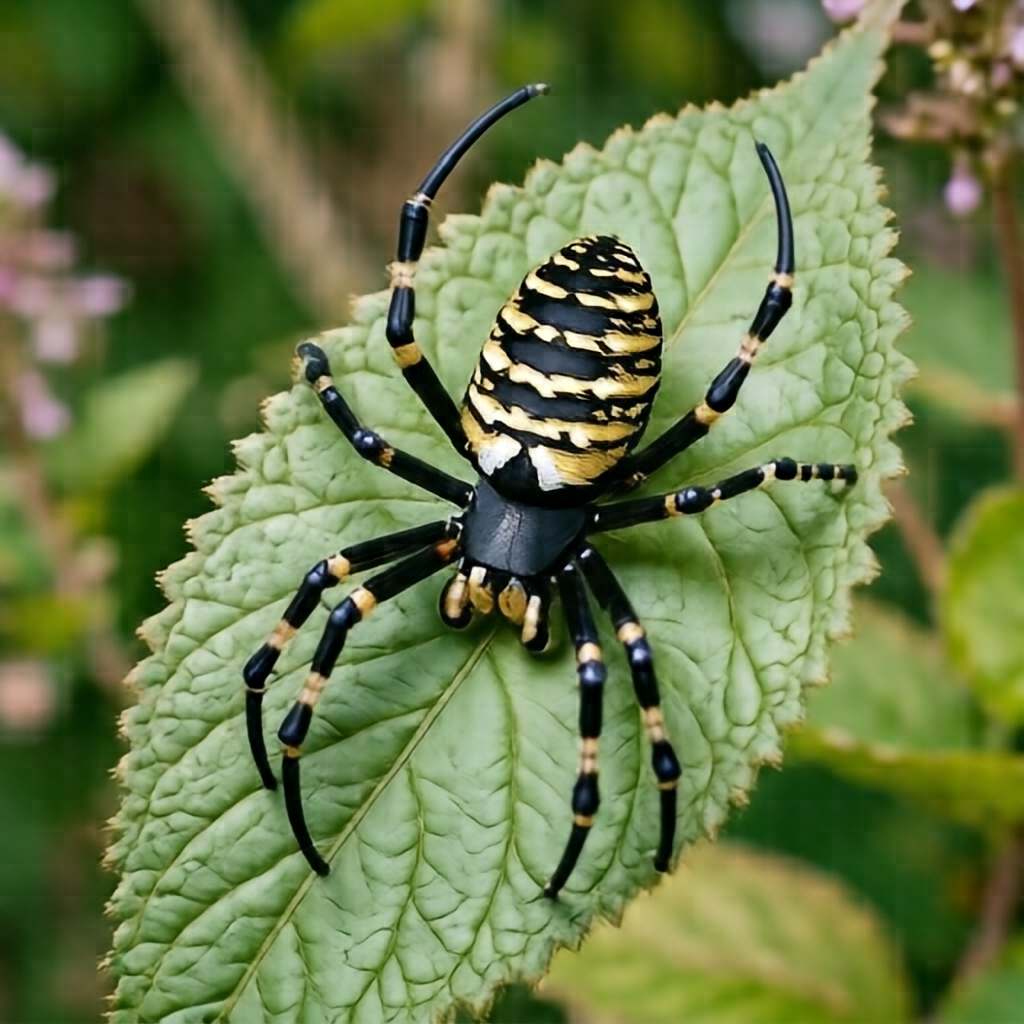}} &
        \begin{tabular}{@{}l@{}}
            \promptboxlarge{Background-description prompt (Qwen3-VL)}{%
                \texttt{The background consists of out-of-focus green foliage with soft, blurred patches of pinkish-purple flowers visible in the upper right and left corners, creating a natural, garden-like setting with shallow depth of field.}%
            } \\[4pt]
            \promptboxlarge{Style-description prompt (Qwen3-VL)}{%
                \texttt{The image presents a vivid, sharply focused macro view of a strikingly patterned spider resting on a textured green leaf, with a soft, blurred natural background that enhances the subject's bold black-and-yellow contrast and creates a sense of quiet, detailed observation in a lush, living environment.}%
            }
        \end{tabular} \\
        \midrule
        \raisebox{-0.5\height}{\includegraphics[width=0.215\textwidth]{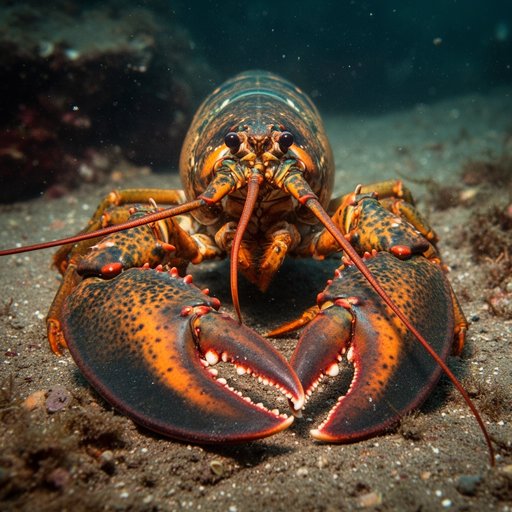}} &
        \begin{tabular}{@{}l@{}}
            \promptboxlarge{Background-description prompt (InternVL3.5)}{%
                \texttt{The background of the image is a sandy seabed with some scattered debris and small particles, giving it a natural underwater appearance. The area behind the main subject is slightly blurred, emphasizing the foreground.}%
            } \\[4pt]
            \promptboxlarge{Style-description prompt (InternVL3.5)}{%
                \texttt{The image has a vibrant and intense visual style, with the bright colors of the lobster contrasting against the dark, murky background, creating a dramatic and somewhat menacing atmosphere. The close-up perspective adds a sense of immediacy and detail, drawing the viewer into the scene.}%
            }
        \end{tabular} \\
        \midrule
        \raisebox{-0.5\height}{\includegraphics[width=0.215\textwidth]{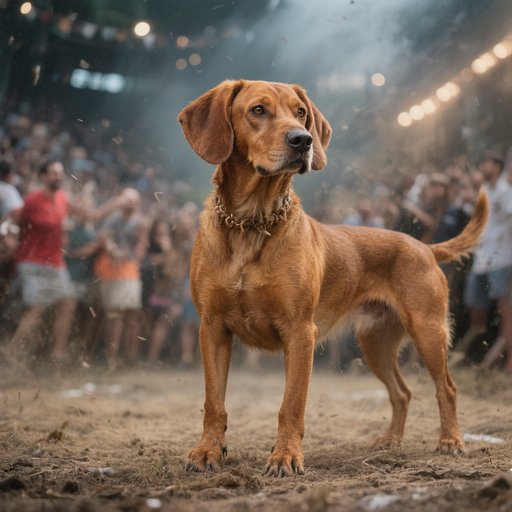}} &
        \begin{tabular}{@{}l@{}}
            \promptboxlarge{Background-description prompt (GLM-4.1V)}{%
                \texttt{The background shows a large crowd of people gathered in what appears to be an outdoor setting with dim lighting and string lights visible. There's a hazy atmosphere with some light particles or dust in the air, suggesting a festival or event environment.}%
            } \\[4pt]
            \promptboxlarge{Style-description prompt (GLM-4.1V)}{%
                \texttt{The image has a cinematic, atmospheric style with a soft focus and warm tones, creating a moody, almost dreamlike quality. The scene feels dynamic and lively, with a sense of energy from the blurred crowd and particles in the air, while the dog stands as a clear focal point with a calm, dignified presence.}%
            }
        \end{tabular} \\
        \bottomrule
    \end{tabular}
    \caption{\textbf{Examples of captions under the background and style prompts.} Captions are reproduced from the saved experimental outputs, with only typographic punctuation normalized.}
    \label{tbl:appx_caption_examples}
\end{table*}

\section{Limitations and Future Work}\label{sec:appx_limitations}

Our study is intentionally scoped to four open-source frontier \ac{t2i} models, two English prompt families, and one default generation configuration per model. The resulting accuracies therefore establish strong closed-set distinguishability within this roster, but do not guarantee attribution to proprietary models or future checkpoints. Moreover, each source label jointly reflects model weights, architecture, training data, sampler, decoder, and any model-specific post-processing; our analyses reveal product-level visual idiosyncrasies but do not causally localize them to a particular component.

Future work should disentangle these factors through controlled comparisons across checkpoints, model families, samplers, guidance scales, inference steps, resolutions, and shared decoding pipelines, while extending evaluation to open-set and continually updated model rosters. In addition, the mid-level maps and high-level descriptions are estimated by pretrained segmentation, depth, inverse-rendering, and captioning models and may inherit their biases or errors. Replication with alternative extractors, human annotations, and direct interventions on the identified visual properties would help distinguish stable generator behavior from measurement artifacts and clarify which idiosyncrasies causally support human and machine source classification.

\end{document}

%% file: config.tex
\usepackage{amsmath}
\usepackage{amssymb}
\usepackage{booktabs}
\usepackage{multirow}
\usepackage{subcaption}
\usepackage{acronym}
\usepackage[capitalise,nosort]{cleveref}

\crefname{section}{Sec.}{Secs.}
\Crefname{section}{Section}{Sections}
\crefname{subsection}{Sec.}{Secs.}
\Crefname{subsection}{Section}{Sections}
\crefname{appendix}{Appx.}{Appxs.}
\Crefname{appendix}{Appendix}{Appendices}
\crefname{subappendix}{Appx.}{Appxs.}
\Crefname{subappendix}{Appendix}{Appendices}
\crefname{figure}{Fig.}{Figs.}
\Crefname{figure}{Figure}{Figures}
\crefname{table}{Tab.}{Tabs.}
\Crefname{table}{Table}{Tables}
\crefname{equation}{Eq.}{Eqs.}
\Crefname{equation}{Equation}{Equations}

\acrodef{t2i}[T2I]{Text-to-Image}
\acrodef{llm}[LLM]{Large Language Model}
\acrodef{vlm}[VLM]{Vision-Language Model}
\acrodef{mlp}[MLP]{Multi-Layer Perceptron}
\acrodef{cnn}[CNN]{Convolutional Neural Network}
\acrodef{vae}[VAE]{Variational Autoencoder}
\acrodef{tfidf}[TF-IDF]{Term Frequency-Inverse Document Frequency}
\acrodef{glcm}[GLCM]{Gray-Level Co-occurrence Matrix}
\acrodef{lbp}[LBP]{Local Binary Pattern}
\acrodef{hog}[HOG]{Histogram of Oriented Gradients}
\acrodef{fft}[FFT]{Fast Fourier Transform}
\acrodef{id}[ID]{in-distribution}
\acrodef{ood}[OOD]{out-of-distribution}

\acrodef{hidream}[HiDream]{HiDream-O1-Dev}
\acrodef{flux}[FLUX.2]{FLUX.2-dev}
\acrodef{hunyuan}[HunyuanImage]{HunyuanImage-3.0}
\acrodef{qwenimage}[Qwen-Image]{Qwen-Image-2512}

\newcommand{\cmark}{\ensuremath{\checkmark}}

\newcommand{\hunyuanstack}{\shortstack{Hunyuan-\\Image}}
\newcommand{\qwenimagestack}{\shortstack{Qwen-\\Image}}
\def\onedot{.}
\def\eg{\emph{e.g}\onedot}
\def\ie{\emph{i.e}\onedot}

\newcommand{\promptbox}[2]{%
    \begin{center}
        \fcolorbox{black!18}{black!3}{%
            \parbox{\dimexpr\columnwidth-2\fboxsep-2\fboxrule\relax}{%
                \small\raggedright
                \textbf{#1}\par
                #2\par
            }%
        }%
    \end{center}
}

\newcommand{\promptboxlarge}[2]{%
    \fcolorbox{black!18}{black!3}{%
        \parbox{\dimexpr1.5\columnwidth-2\fboxsep-2\fboxrule\relax}{%
            \small\raggedright
            \textbf{#1}\par
            #2\par
        }%
    }%
}